\documentclass[11pt]{article}

\usepackage[final]{acl}

\usepackage{times}
\usepackage{latexsym}

\usepackage[T1]{fontenc}

\usepackage[utf8]{inputenc}

\usepackage{microtype}

\usepackage{inconsolata}

\usepackage{graphicx}
\usepackage{xcolor}
\usepackage{amssymb}
\usepackage{tikz}
\usepackage{booktabs}   
\usepackage{multirow}   
\usepackage[table]{xcolor} 
\usepackage{colortbl} 
\definecolor{graybg}{rgb}{0.95, 0.95, 0.95} 
\usepackage{pifont} 
\usepackage{booktabs}
\usepackage{multirow}
\usepackage{colortbl}
\usepackage{xcolor}
\usepackage[most]{tcolorbox} 

\definecolor{pOrange}{HTML}{D35400} 
\definecolor{pBlue}{HTML}{1F618D}   
\definecolor{pRed}{HTML}{C0392B}    
\definecolor{bgGray}{HTML}{F7F9F9}  

\newcommand{\promptkey}[1]{\textbf{\textcolor{pOrange}{#1}}}
\newcommand{\promptuser}{\textbf{\textcolor{pBlue}{USER}}}
\newcommand{\promptvar}[1]{\textcolor{pRed}{\{#1\}}}

\usepackage{enumitem}

\definecolor{humangreen}{HTML}{E6F4EA}

\title{Do MLLMs Understand Pointing? Benchmarking and Enhancing Referential Reasoning in Egocentric Vision}

\author{
  Chentao Li\textsuperscript{1},
  Zirui Gao\textsuperscript{1},
  Mingze Gao\textsuperscript{2}, 
  \textbf{Yinglian Ren\textsuperscript{1}},
  \textbf{Jianjiang Feng\textsuperscript{1}\thanks{\ \ Corresponding author.}},
  \textbf{Jie Zhou\textsuperscript{1}} \\
  \textsuperscript{1}Department of Automation, Tsinghua University \\
  \textsuperscript{2}Academy of Art \& Design, Tsinghua University \\
  \texttt{lict23@mails.tsinghua.edu.cn},
  \texttt{jfeng@tsinghua.edu.cn}
}

\begin{document}
\maketitle
\begin{abstract}

Egocentric AI agents, such as smart glasses, rely on pointing gestures to resolve referential ambiguities in natural language commands. However, despite advancements in Multimodal Large Language Models (MLLMs), current systems often fail to precisely ground the spatial semantics of pointing. Instead, they rely on spurious correlations with visual proximity or object saliency—a phenomenon we term ``Referential Hallucination.'' To address this gap, we introduce \texttt{EgoPoint-Bench}, a comprehensive question-answering benchmark designed to evaluate and enhance multimodal pointing reasoning in egocentric views. Comprising over 11k high-fidelity simulated and real-world samples, the benchmark spans five evaluation dimensions and three levels of referential complexity. Extensive experiments demonstrate that while state-of-the-art proprietary and open-source models struggle with egocentric pointing, models fine-tuned on our synthetic data achieve significant performance gains and robust Sim-to-Real generalization. This work highlights the importance of spatially-aware supervision and offers a scalable path toward precise egocentric AI assistants. The project website is available at \url{https://guyyyug.github.io/EgoPoint-Bench/}.
\end{abstract}

\section{Introduction}

\begin{figure}[t]
  \includegraphics[width=\columnwidth]{image/teaser.pdf}
\caption{Spatial ambiguity in egocentric pointing.
Two examples where current VLMs (e.g., Gemini 3, Qwen3-VL) fail to recognize the target spatially aligned with the pointing gesture. 
This highlights a critical gap in fine-grained 3D spatial reasoning. Note that neither bboxes nor rays were in the model inputs.}
  \label{fig:teaser}
\end{figure}

Egocentric Vision AI agents, particularly intelligent assistants integrated into wearable devices such as smart glasses, are fundamentally reshaping the paradigms of Augmented Reality and Human-Computer Interaction ~\cite{li2025challenges}. By perceiving the physical world through the user's perspective, these systems aim to provide precise, context-aware Question Answering (QA) services. In such naturalistic interaction scenarios, users exhibit a strong preference for minimalistic spoken commands. These utterances often blend explicit object descriptions with highly ambiguous deictic expressions (e.g., ``\textit{How do I use this?}'' or ``\textit{How is the stuff over there?}''). When retrieving information from complex visual scenes, relying solely on unimodal language is often insufficient to resolve such referential ambiguity. Conversely, pointing gestures—instinctual and high-frequency actions in human communication—have been empirically proven to significantly enhance referential clarity and reduce the requisite length of natural language instructions~\cite{mane2025ges3vig,chen2021yourefit}. Consequently, endowing multimodal models with the capability to precisely comprehend ``egocentric pointing'' is critical for egocentric AI agents.

Despite the remarkable semantic understanding demonstrated by Multimodal Large Language Models (MLLMs) in general image captioning and QA tasks~\cite{gpt4o, llava}, our investigation reveals a critical deficiency in spatial reasoning when adapting current state-of-the-art models to egocentric pointing QA. Specifically, as depicted in Fig. \ref{fig:teaser}, instead of tracing the precise geometric projection of the pointing finger, models frequently fixate on objects proximal to the hand or visually salient entities, leading to \textit{referential hallucination}. This indicates that these models fail to grasp the intrinsic spatial mechanism of ``pointing'', relying instead on spurious correlations based on visual proximity.

A critical bottleneck is the scarcity of high-quality, unambiguous data aligned within the ``Vision-Language-Space''. While visual grounding is well-studied, benchmarks like RefCOCO~\cite{kazemzadeh2014referitgame} and Visual Genome~\cite{krishna2017visual} rely on third-person internet imagery, lacking the wide-angle nature of egocentric vision. Conversely, large egocentric datasets like Ego4D~\cite{grauman2022ego4d} and EPIC-KITCHENS~\cite{damen2022rescaling} prioritize action recognition or hand-object interactions~\cite{liu2022hoi4d}, missing dense QA annotations that capture ``pointing-object'' geometry. Without this spatially-aware supervision, MLLMs fail to separate hand appearance from spatial pointing intent, hindering deictic referencing performance.

To address this challenge, we propose \texttt{EgoPoint-Bench}, a benchmark designed to systematically evaluate and enhance multi-modal spatial reasoning in egocentric views. 
Our construction process balances data scale with realism through two complementary phases.
In the simulation phase, we introduce a physics-based synthesis pipeline leveraging ray-casting to generate noise-free pointing labels in 3D environments; in the real-world phase, we collect real-scenario data to validate practical applicability. For QA construction, we implemented a hybrid ``machine-generation, human-verification'' pipeline to ensure rigorous standards. Crucially, to capture interaction diversity and enable fine-grained assessment, we incorporated three referring language patterns ranging from explicit descriptions to implicit instructions, and structured the benchmark across five core capability dimensions. In total, the dataset comprises 10,567 high-fidelity simulation QA pairs and 1,162 real-world samples.

To evaluate generalization, we employed a hybrid test set combining held-out simulation data (in-domain) and real-world data (zero-shot cross-domain). We benchmarked open-source (e.g., Qwen3-VL) and proprietary models (e.g., GPT-5), followed by LoRA fine-tuning on simulation data. The fine-tuned models consistently outperform their direct-inference baselines and demonstrate effective sim-to-real generalization on real-world test sets. These results validate the efficacy of high-quality synthetic data and highlight the scarcity of egocentric pointing examples in current foundation models.
The main contributions of this paper are summarized as follows:
\begin{itemize}[leftmargin=*]
    \item We propose \texttt{EgoPoint-Bench}, a novel benchmark designed to evaluate multi-modal spatial reasoning in egocentric views. Our extensive benchmarking reveals that current state-of-the-art MLLMs significantly lack the capability to understand fine-grained pointing gestures in first-person scenarios.
    
    \item We develop a physics-driven data generation pipeline that ensures both geometric precision and linguistic diversity. By leveraging ray-casting in simulation and incorporating hierarchical referring patterns (from explicit descriptions to implicit instructions), we construct a high-quality dataset containing over 11k pairs across simulation and real-world domains.
    
    \item We demonstrate effective sim-to-real generalization. Models fine-tuned on our high-fidelity synthetic data achieve consistent improvements on real-world test sets, validating the potential of synthetic data for addressing data scarcity in egocentric interaction.
\end{itemize}

\section{Related Work}

To contextualize our contributions, we compare \texttt{EgoPoint-Bench} with representative benchmarks in visual grounding, embodied perception, and pointing-based interaction (see Table~\ref{tab:dataset_comparison}).

\begin{table*}[t]
\centering
\caption{Comparison with existing vision-language and embodied cognition datasets. Unlike previous benchmarks that inherently rely on third-person static views, algorithmically synthetic avatars, or artificial visual prompts (such as bounding boxes drawn on images), \texttt{EgoPoint-Bench} uniquely unifies true egocentric vision with real-world natural hand pointing mechanics. It overcomes the spatial constraints of prior works by supporting diverse question types and multi-level linguistic granularity for robust MLLM evaluation.
\textbf{R}: Real-world data, \textbf{S}: Synthetic data.}
\label{tab:dataset_comparison}
\resizebox{\textwidth}{!}{%
\begin{tabular}{l|c|c|c|c|l|c}
\toprule
\textbf{Dataset} & \textbf{Egocentric} & \textbf{Scenes} & \textbf{Natural Pointing} & \textbf{Task} & \textbf{Annotation Granularity} & \textbf{Size} \\
\midrule
RefCOCOg~\cite{mao2016generation} & \ding{55} & \textbf{R} & \ding{55} & Grounding & Image + BBox + Text & 26k imgs \\
ScanRefer~\cite{chen2020scanrefer} & \ding{55} & \textbf{R} & \ding{55} & 3D Grounding & 3D BBox + Text & 51.5k expr. \\
YouRefIt~\cite{chen2021yourefit} & \ding{55} & \textbf{R} & \ding{51} & Grounding & BBox + Gesture + Text & 4.2k clips \\
Ego4D~\cite{grauman2022ego4d} & \ding{51} & \textbf{R} & \ding{55} & Forecasting & Activity Labels & 3.6k hrs \\
Ges3ViG~\cite{mane2025ges3vig} & \ding{55} & \textbf{S} & \ding{55} & 3D Grounding & 3D BBox + Synth. Gesture & 35k samples \\
EOC-Bench~\cite{yuan2025eocbench} & \ding{51} & \textbf{R} & \ding{55} & QA & Temporal QA + Visual Prompts & 3.3k QAs \\
ECBench~\cite{dang2025ecbench} & \ding{51} & \textbf{R} & \ding{55} & QA & Cognitive QA & 4.3k QAs \\
\midrule
\texttt{EgoPoint-Bench (Ours)} & \textbf{\ding{51}} & \textbf{R+S} & \textbf{\ding{51}} & \textbf{QA/Grounding} & \textbf{Image + Name + BBox + QA} & \textbf{11.7k QAs} \\
\bottomrule
\end{tabular}%
}
\end{table*}

\subsection{From Explicit Grounding to Semantic Underspecification}
Foundational visual grounding benchmarks, ranging from 2D~\cite{mao2016generation,krishna2017visual} to 3D~\cite{chen2020scanrefer,achlioptas2020referit3d} and robotic settings~\cite{qi2020reverie}, rely predominantly on third-person views and explicit, exhaustive linguistic descriptions. However, natural human communication frequently employs \textit{semantic underspecification} and \textit{exophora}---using deictic pronouns like ``this'' or ``that'' whose meanings are entirely reliant on the external visual or gestural context. 

\subsection{Egocentric Reasoning and Perception}
Large-scale datasets like Ego4D~\cite{grauman2022ego4d} and EPIC-KITCHENS~\cite{damen2018scaling} capture rich first-person activities but focus primarily on passive observation (e.g., action recognition). Recent findings emphasize that Vision-Language Models (VLMs) fundamentally struggle with egocentric spatial reasoning, especially when tracking objects across temporal shifts and disjoint frames. 
RefEgo~\cite{kurita2023refego} provides language grounding for egocentric video but uses text-only referring expressions and does not incorporate natural gesture signals.
While recent benchmarks like EOC-Bench~\cite{yuan2025eocbench} introduce open-ended QA to egocentric videos, they rely on artificial visual prompts. This reliance creates a significant domain gap for real-world Augmented Reality (AR) applications, where systems must interpret unaugmented, dynamic user cues.

\subsection{Pointing-driven Disambiguation}
To enable pointing-driven interaction, Ges3ViG~\cite{mane2025ges3vig} introduces 3D directional gestures through synthesized avatars; however, it focuses on object localization within 3D scenes rather than complex question-answering (QA) and lacks validation on real-world kinematics. While COSM2IC~\cite{weerakoon2022cosm2ic} achieves deictic interaction using virtual environments, it is limited by a lack of diversity in both object categories and scene types. In contrast, \texttt{EgoPoint-Bench} integrates high-fidelity synthetic and real-world data. We shift linguistic inputs from explicit descriptions (e.g., ``the object I point at'') to implicit deictics (e.g., ``this''), evaluating MLLMs' pointing comprehension across diverse semantic dimensions.

\section{EgoPoint-Bench}
\begin{figure*}[t]
\centering
  \includegraphics[width=\linewidth]{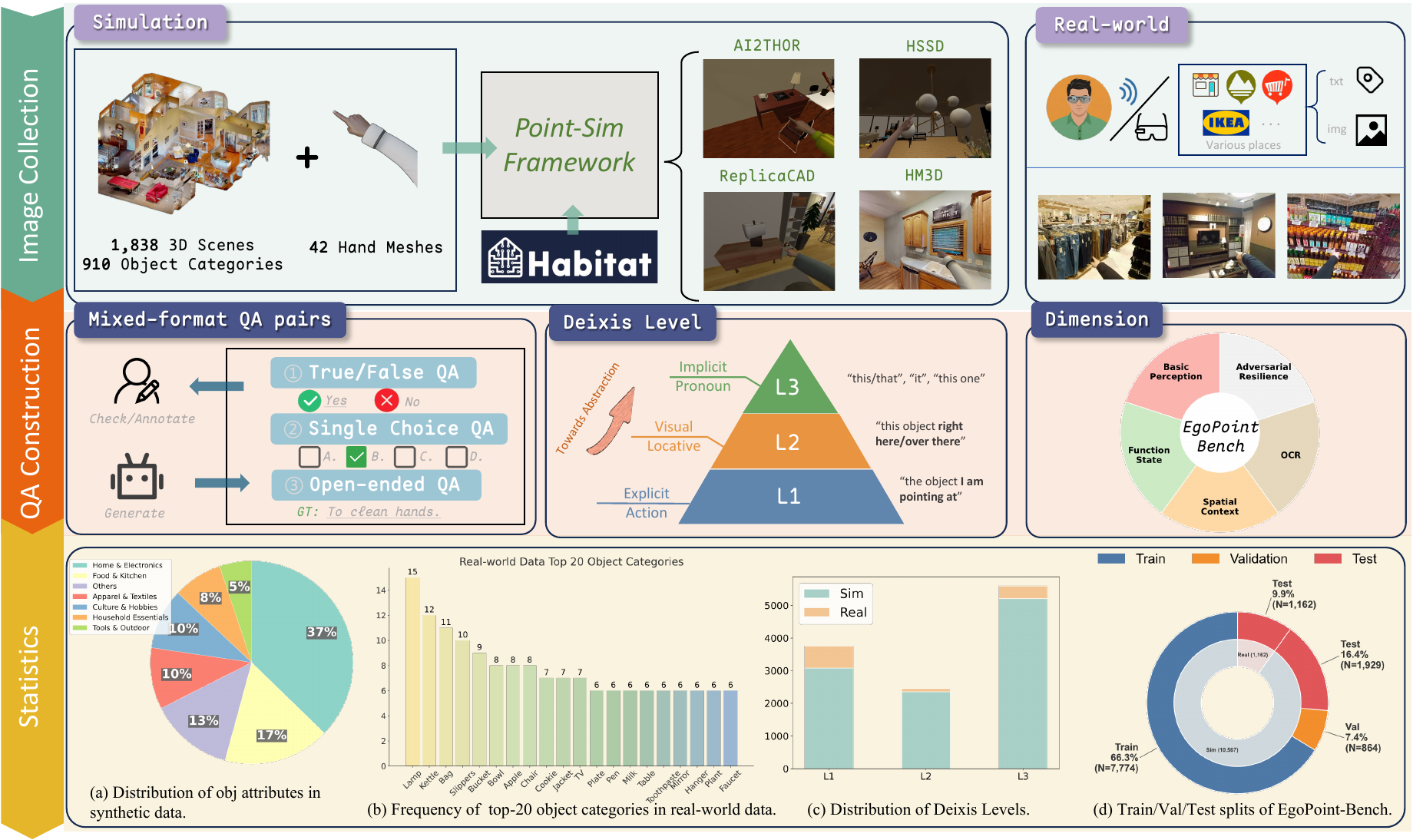}
  \caption{Overview of \texttt{EgoPoint-Bench}.
Top: We construct the dataset using a scalable simulation pipeline (\textit{Point-Sim}) alongside real-world collection to ensure visual diversity. 
Middle: The QA generation process spans five capability dimensions (Basic Perception, Function \& State, Spatial Context, OCR, and Adversarial Resilience) and incorporates a hierarchical deixis level taxonomy (L1: Explicit Action, L2: Visual Locative, L3: Implicit Pronoun), challenging models to resolve referential ambiguity based on finger-pointing gestures. 
Bottom: Detailed statistics showing object attributes, category frequency, and data distribution.}
\label{fig:overview}
\end{figure*}

\subsection{Overview}
As shown in Fig. \ref{fig:overview}, we propose \texttt{EgoPoint-Bench}, a multimodal question-answering benchmark focused on first-person pointing gestures. It is designed to quantitatively evaluate the understanding and reasoning capabilities of MLLMs regarding pointing gestures and referring language in egocentric visual perception. Given the scarcity of labeled data in this domain, we employ a dual-source data construction strategy combining simulation and real-world data. On one hand, we introduce the \textbf{Point-Sim} fully automated simulation framework, which utilizes 42 hand models to generate 10,567 synthetic samples across 1,838 high-fidelity 3D scenes (sourced from Ai2-THOR \cite{ai2thor,procthor}, HSSD \cite{khanna2023hssd}, ReplicaCAD \cite{szot2021habitat}, and HM3D \cite{ramakrishnan2021hm3d}). On the other hand, to enhance the realistic diversity of the dataset, we collected 1,162 samples featuring natural pointing interactions in diverse real-world environments. Furthermore, the benchmark covers five core dimensions and includes three question types—multiple-choice, true/false, and open-ended questions—with established standard splits for training, validation, and testing.

\subsection{Image Collection}
\subsubsection{Point-Sim Simulation Framework}
\label{sec:point_sim}
To synthesize diverse and high-fidelity scene-object pairs, we utilized the Habitat-Sim 3.0 simulator~\cite{puig2023habitat3} and integrated static environments sourced from the AI2-THOR, HSSD, ReplicaCAD, and HM3D datasets. Specifically, we acquired high-quality 3D arm-hand models from ArtStation \cite{artstation} and leveraged the Blender package~\cite{blender2018} to manipulate parameters---such as joint articulation and scaling---thereby introducing structural diversity into the generated pointing gestures. Furthermore, we applied textures representing 3 distinct skin tones and 7 clothing styles across both left and right hands, resulting in a total of 42 unique pointing models.

\paragraph{Simulation Initialization.}
To ensure domain robustness, we initialize the simulation with a diverse set of intrinsic and extrinsic parameters. To replicate the wide-angle optical characteristics of modern smart glasses, the camera's vertical field of view (FOV) is uniformly sampled from $[100^\circ, 115^\circ]$. The agent is modeled with an ocular height $h_{eye} \sim \mathcal{U}(1.45, 1.70)$ meters, equipped with a multi-modal sensor suite capturing aligned RGB, Depth, and Semantic observations. Hand dominance (left/right) is randomized to balance the dataset distribution.

\paragraph{Target-Oriented Spatial Arrangement.}
For a selected target object $O$ centered at $P_{obj} \in \mathbb{R}^3$, we compute the navigable manifold of the scene, represented as a Navigation Mesh (NavMesh)~\cite{mononen2009recast}. We sample a candidate agent position $P_{agent}$ on this manifold within a constrained radius $r_{search}$ (default $\le 3.0$m), conditioned on a minimum collision clearance of $0.4$m. To mitigate scale ambiguity, the sampling distance is dynamically scaled based on the object's volumetric size; this prevents scenarios where the object is either imperceptible or encompasses the entire field of view.

Once $P_{agent}$ is fixed, we orient the agent's camera to face the target. We construct the camera rotation matrix $R_{cam} \in SO(3)$ by aligning the optical axis with the forward vector $\mathbf{f} = (P_{obj} - P_{agent}) / \|P_{obj} - P_{agent}\|$. The rotation is defined compactly as:
\begin{equation}
    R_{cam} = \left[ \frac{\mathbf{f} \times \mathbf{u}_w}{\|\mathbf{f} \times \mathbf{u}_w\|}, \;\; \frac{(\mathbf{f} \times \mathbf{u}_w) \times \mathbf{f}}{\|\mathbf{f} \times \mathbf{u}_w\|}, \;\; -\mathbf{f} \right]^\top
\end{equation}
where $\mathbf{u}_w$ is the global up vector.

\paragraph{Kinematic Hand Alignment.}
We instantiate the hand model within the lower visual field of the camera. The core objective is to align the index finger's direction vector with the line of sight to the object. Let $\mathbf{u}_{rest}$ denote the normalized initial directional vector of the index finger and $\mathbf{u}_{target}$ be the normalized vector pointing from the hand to the object. We compute the minimal rotation $R_{hand}$ via \textit{Rodrigues' rotation formula}. The rotation is parameterized by the unit rotation axis $\mathbf{k} = \frac{\mathbf{u}_{rest} \times \mathbf{u}_{target}}{\|\mathbf{u}_{rest} \times \mathbf{u}_{target}\|}$ and angle $\theta = \arccos(\mathbf{u}_{rest} \cdot \mathbf{u}_{target})$:
\begin{equation}
    R_{hand} = I + [\mathbf{k}]_\times \sin\theta + [\mathbf{k}]_\times^2 (1 - \cos\theta)
\end{equation}
where $[\mathbf{k}]_\times$ denotes the skew-symmetric matrix of $\mathbf{k}$. Subsequently, to simulate realistic human pointing behavior, we apply small stochastic perturbations to the pitch and yaw of the computed camera orientation.

\begin{figure}[t]
  \includegraphics[width=\columnwidth]{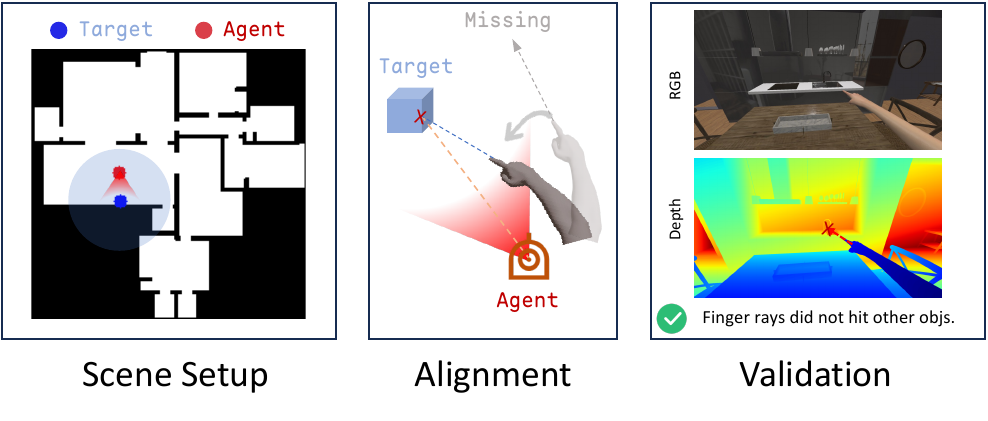}
  \caption{Point-sim Simulation Framework.}
  \label{fig:sim-datas}
\end{figure}

\paragraph{Validation and Data Format.}
We enforce a validity check by casting a ray from the index finger tip toward $P_{obj}$. An instance is discarded if the ray intersects with any obstacle before reaching the target. The pipeline explicitly exports a comprehensive data tuple $\mathcal{D} = \{I_{rgb}, I_{depth}, I_{sem}, \mathbf{b}_{obj}, P_{2D}, y_{id}\}$, containing the images, 2D bounding boxes, projected coordinates, and semantic identifiers. This pipeline is generalized to support any scene compatible with Habitat-Sim.

\subsubsection{Real-world Data Collection}
We recruited eight volunteers equipped with MLVision smart glasses~\cite{mlvision_web} to collect data on objects of interest in diverse real-world environments. The data collection scenarios spanned a broad spectrum of settings, including but not limited to indoor places like furniture stores, convenience stores, and apartments, as well as outdoor locations such as shopping malls, zoos, and streets. Participants were instructed to record a video whenever they encountered an object of interest, explicitly pointing at the target while verbally stating its name to serve as the ground truth and posing a relevant description or question. In total, 1,162 valid image frames were curated from the collected footage (see Appendix \ref{subsec:real_world} for details).

\subsection{Capability Taxonomy}

Inspired by canonical multimodal benchmarks like MMBench~\cite{liu2024mmbench} and MME~\cite{fu2025mmecomprehensiveevaluationbenchmark}, we design a five-dimensional taxonomy to comprehensively evaluate MLLMs within first-person pointing interactions. This framework is structured to bridge the gap between low-level perception and high-level robust reasoning:

\begin{itemize}[leftmargin=*, nosep]
    \item \textbf{Basic Perception (BP):} Identifies fundamental attributes (category, color, texture) and visual distinctiveness for gesture alignment.
    \item \textbf{Function \& State (FS):} Infers semantic properties (e.g., edibility, operability) and dynamic functional states.
    \item \textbf{Spatial Context (SC):} Perceives egocentric spatial relationships, including localization, scene compatibility, and reachability.
    \item \textbf{OCR:} Extracts textual info from targets, such as brand names, slogans, and instructions.
    \item \textbf{Adversarial Resilience (AR):} Maintains reliability against adversarial inputs like counterfactuals, fallacies, and void references.
\end{itemize}

\begin{table*}[t]
\centering
\definecolor{graybg}{rgb}{0.94, 0.94, 0.94}
\definecolor{lightorange}{rgb}{1.0, 0.92, 0.8}
\definecolor{lightblue}{rgb}{0.85, 0.93, 1.0}
\definecolor{darkgreen}{rgb}{0.0, 0.5, 0.0}
\definecolor{lightgreen}{rgb}{0.88, 1.0, 0.88}

\setlength{\aboverulesep}{0pt}
\setlength{\belowrulesep}{0pt}
\renewcommand{\arraystretch}{1}

\caption{Main results on real-world and simulation testsets. We highlight the \colorbox{lightblue}{best Direct results} in blue and the \colorbox{lightorange}{best LoRA results} in orange. The Gain column shows the improvement of LoRA over Direct.}
\label{tab:main_results_final}

\resizebox{\textwidth}{!}{%
\begin{tabular}{l|c|ccccc|c|ccccc|c|cc}
\toprule
\multirow{2}{*}{\textbf{Model}} & \multirow{2}{*}{\textbf{Method}} & \multicolumn{6}{c|}{\textbf{Simulation testset}} & \multicolumn{6}{c|}{\textbf{Real-world testset}} & \multicolumn{2}{c}{\textbf{Overall}} \\
  & & \textbf{BP} & \textbf{FS} & \textbf{SC} & \textbf{OCR} & \textbf{AR} & \textbf{Mean} & \textbf{BP} & \textbf{FS} & \textbf{SC} & \textbf{OCR} & \textbf{AR} & \textbf{Mean} & \textbf{Avg.} & \textbf{Gain} \\
\midrule

Random & - & 27.95 & 26.83 & 38.89 & 43.24 & 52.17 & 31.14 & 25.19 & 22.74 & 37.30 & 26.32 & 45.76 & 28.94 & 30.24 & - \\
\rowcolor{lightgreen} Human & -& 91.86 & 97.14 & 100 & 93.33 & 100 & 95.80& 96.24 & 98.04 & 96.39 & 95.65 & 89.09 & 96.00 & 95.90  & -  \\
\midrule
\rowcolor{graybg} \multicolumn{16}{c}{\textit{Closed-source Models}} \\
\midrule
Gemini 3 Pro   & Direct & 52.47 & 51.39 & \cellcolor{lightblue}70.47 & 74.85 & 57.16 & 56.44 & 66.63 & \cellcolor{lightblue}75.44 & 79.06 & \cellcolor{lightblue}83.28 & 60.16 & \cellcolor{lightblue}72.00 & 62.29 & - \\
Gemini 3 Flash & Direct & 54.39 & 53.33 & 66.58 & 73.64 & 58.39 & 57.21 & \cellcolor{lightblue}67.04 & 73.98 & 78.89 & 80.90 & 63.02 & 71.84 & \cellcolor{lightblue}62.71 & - \\
GPT-5.2 Instant  & Direct & 54.14 & 49.81 & 66.14 & 75.45 & 50.88 & 54.80 & 55.31 & 67.49 & \cellcolor{lightblue}81.62 & 69.55 & \cellcolor{lightblue}71.27 & 66.76 & 59.29 & - \\
GPT-5 mini       & Direct & \cellcolor{lightblue}59.96 & \cellcolor{lightblue}58.22 & 67.65 & 68.79 & 36.09 & 57.66 & 52.81 & 66.73 & 67.32 & 66.27 & 52.38 & 60.57 & 58.75 & - \\
\midrule

\rowcolor{graybg} \multicolumn{16}{c}{\textit{Open-source Models (Direct vs. LoRA)}} \\
\midrule

\multirow{2}{*}{LLaVA-1.5-7B} & Direct & 50.83 & 46.89 & 54.86 & 50.91 & 41.92 & 48.82 & 36.48 & 45.85 & 62.13 & 22.69 & 69.37 & 47.19 & 48.21 & - \\
 & LoRA & 76.41 & 72.06 & 60.63 & 66.06 & 86.44 & 73.18 & 37.50 & 56.55 & 64.17 & 33.43 & \cellcolor{lightorange}95.40 & 54.54 & 66.17 & \textcolor{darkgreen}{+17.96} \\
\cmidrule{1-16}

\multirow{2}{*}{LLaVA-NeXT-7B} & Direct & 47.42 & 45.42 & 55.92 & 53.33 & 46.59 & 48.17 & 31.68 & 51.75 & 60.09 & 39.40 & 56.19 & 46.44 & 47.52 & - \\
 & LoRA & 80.39 & 80.86 & 79.56 & 72.42 & 86.13 & 80.93 & 40.10 & 66.32 & 71.23 & 40.90 & 90.63 & 59.64 & 72.93 & \cellcolor{lightorange}\textbf{\textcolor{darkgreen}{+25.41}} \\
\cmidrule{1-16}

\multirow{2}{*}{GLM-4.6V-Flash} & Direct & 56.16 & 50.81 & 66.14 & 61.52 & 36.17 & 53.29 & 48.32 & 59.77 & 67.32 & 72.84 & 43.49 & 56.42 & 54.47 & - \\
 & LoRA & 77.16 & 73.28 & 82.01 & 80.00 & 64.21 & 74.86 & 53.88 & 60.70 & 66.55 & 67.16 & 72.70 & 61.26 & 69.74 & \textcolor{darkgreen}{+15.27} \\
\cmidrule{1-16}

\multirow{2}{*}{InternVL3.5-2B} & Direct & 51.97 & 55.14 & 61.50 & 66.97 & 26.05 & 51.74 & 44.85 & 60.47 & 62.55 & 59.40 & 43.65 & 53.73 & 52.49 & - \\
 & LoRA & 71.40 & 75.36 & 76.61 & 78.79 & 81.99 & 75.43 & 46.33 & 64.04 & 71.83 & 57.31 & 89.68 & 62.03 & 70.39 & \textcolor{darkgreen}{+17.90} \\
\cmidrule{1-16}

\multirow{2}{*}{InternVL3.5-8B} & Direct & 52.86 & 52.50 & 63.51 & 66.36 & 35.63 & 52.62 & 50.05 & 60.88 & 63.32 & 68.96 & 50.79 & 57.09 & 54.30 & - \\
 & LoRA & 74.60 & 77.81 & 82.76 & 78.79 & 86.21 & 78.86 & 50.56 & 69.88 & 74.47 & 63.88 & 90.00 & 66.13 & 74.07 & \textcolor{darkgreen}{+19.77} \\
\cmidrule{1-16}

\multirow{2}{*}{InternVL3.5-14B} & Direct & 46.79 & 51.14 & 62.07 & 71.52 & 33.56 & 49.99 & 47.76 & 65.09 & 72.51 & 65.07 & 45.24 & 58.59 & 53.23 & - \\
 & LoRA & 75.99 & 76.00 & 83.01 & 76.36 & \cellcolor{lightorange}86.51 & 78.59 & 54.03 & \cellcolor{lightorange}73.10 & 80.26 & 68.66 & 82.86 & 68.92 & 74.95 & \textcolor{darkgreen}{+21.72} \\
\cmidrule{1-16}

\multirow{2}{*}{Qwen3-VL-8B} & Direct & 57.55 & 54.00 & 70.34 & 77.58 & 52.11 & \cellcolor{lightblue}58.29 & 47.81 & 58.42 & 74.55 & 68.96 & 53.17 & 58.14 & 58.23 & - \\
 & LoRA & \cellcolor{lightorange}81.31 & 80.92 & 80.56 & \cellcolor{lightorange}84.24 & 82.91 & 81.36 & 60.36 & 72.28 & \cellcolor{lightorange}81.96 & 71.94 & 88.57 & \cellcolor{lightorange}71.96 & 77.83 & \textcolor{darkgreen}{+19.60} \\
\cmidrule{1-16}

\multirow{2}{*}{Qwen3-VL-32B} & Direct & 56.52 & 53.75 & 65.64 & \cellcolor{lightblue}79.39 & \cellcolor{lightblue}60.23 & 58.28 & 56.38 & 65.03 & 76.09 & 79.70 & 56.83 & 64.30 & 60.54 & - \\
 & LoRA & 80.75 & \cellcolor{lightorange}82.50 & \cellcolor{lightorange}83.39 & 83.03 & 82.84 & \cellcolor{lightorange}82.20 & \cellcolor{lightorange}62.09 & 71.35 & \cellcolor{lightorange}81.96 & \cellcolor{lightorange}73.43 & 83.81 & 71.84 & \cellcolor{lightorange}78.30 & \textcolor{darkgreen}{+17.76} \\

\bottomrule
\end{tabular}%
}
\end{table*}

\subsection{QA Pair Construction}
\label{sec:qa_construction}

For comprehensive deictic evaluation, our dataset employs a hierarchical taxonomy and hybrid question format.

\noindent\textbf{Hierarchical Deixis Taxonomy.} We design three levels of deixis to cover the broadest possible semantic range of referential inquiries:
L1 (Explicit Action) describes the gesture directly (e.g., ``the object I am pointing at'');
L2 (Visual Locative) implies spatial proximity (e.g., ``that thing over there''); and
L3 (Implicit Pronoun) relies purely on visual context (e.g., ``this'').

\noindent\textbf{Task Formulation.} To balance ecological validity with objective evaluation, we adopt diverse question formats. We incorporate Open-ended questions to reflect the natural, unrestricted nature of human inquiry. However, to ensure a fair, consistent, and automated testing benchmark, we also construct True/False and Single-Choice Questions. This hybrid composition retains the semantic complexity of realistic user intent while facilitating rigorous quantitative comparison.

\noindent\textbf{Human-Machine Collaborative Data Curation.} To ensure both diversity and scalability, we established a collaborative data generation pipeline. For the simulation subset, we leveraged a generative model to synthesize QA pairs, thereby mitigating the rigidity of fixed templates and expanding the dimensionality of potential questions~\cite{llava}. To prevent model hallucinations—specifically the misidentification of pointed-at objects—we implemented a visual prompting strategy~\cite{yang2023set}: ground-truth bounding boxes were rendered directly onto the input images to explicitly guide the model's focus. Furthermore, ground-truth category labels and attributes were injected into text prompts to ensure context-aware responses. We validated the fidelity of this automated pipeline through a manual inspection of the test set, identifying and correcting a minimal 3\% error rate. The real-world dataset followed a rigorous human-in-the-loop workflow. 
Annotators labeled the bounding boxes of target objects based on raw open-ended descriptions or questions. Additionally, they provided factual answers and underwent strict cross-verification.

\subsection{Dataset Statistics}
\label{sec:stats}

\texttt{EgoPoint-Bench} comprises 10,567 simulation and 1,162 real-world QA pairs, with an average question length of 9.81 words. The simulation subset is partitioned into 8,638 samples for training/validation (9:1 split) and 1,929 for testing, while the real-world data serves exclusively as a test set. To ensure rigorous evaluation, each \texttt{(scene, object)} tuple in the simulation data appears exactly once. The dataset covers 1,838 unique scenes and 910 object categories. Fig. \ref{fig:overview} presents detailed statistics regarding (a) synthetic object attributes, (b) top-20 real-world object categories, (c) deixis levels, and (d) dataset splits.

\section{Experiments}

\begin{table*}[t]
\centering
\definecolor{graybg}{rgb}{0.94, 0.94, 0.94}
\definecolor{lightorange}{rgb}{1.0, 0.92, 0.8} 
\definecolor{lightblue}{rgb}{0.85, 0.93, 1.0}  
\definecolor{lightgreen}{rgb}{0.88, 1.0, 0.88}

\newcommand{\bestD}[1]{\cellcolor{lightblue}#1}   
\newcommand{\bestL}[1]{\cellcolor{lightorange}#1}  

\setlength{\aboverulesep}{0pt}
\setlength{\belowrulesep}{0pt}
\renewcommand{\arraystretch}{1.1} 

\newcolumntype{Y}{>{\centering\arraybackslash}p{0.9cm}}

\caption{Detailed Breakdown by Question Type. Types: Single-Choice (\small $\mathcal{SCQ}$), True/False(\small $\mathcal{TF}$), Open-Ended questions (\small $\mathcal{OQ}$). Dimensions: Basic Perception (BP), Function \& State (FS), Spatial Context (SC), OCR \& Text (OCR), Adversarial Resilience (AR). \colorbox{lightblue}{Blue} indicates best Direct performance; \colorbox{lightorange}{Orange} indicates best LoRA performance.}
\label{tab:detailed_results}

\resizebox{\textwidth}{!}{%
\begin{tabular}{l|c|YYY|YYY|YYY|YYY|YYY}
\toprule
\multirow{2}{*}{\textbf{Model}} & \multirow{2}{*}{\textbf{Method}} & \multicolumn{3}{c|}{\textbf{BP}} & \multicolumn{3}{c|}{\textbf{FS}} & \multicolumn{3}{c|}{\textbf{SC}} & \multicolumn{3}{c|}{\textbf{OCR}} & \multicolumn{3}{c}{\textbf{AR}} \\
& & \small $\mathcal{SCQ}$ & \small $\mathcal{TF}$ & \small $\mathcal{OQ}$ & \small $\mathcal{SCQ}$ & \small $\mathcal{TF}$ & \small $\mathcal{OQ}$ & \small $\mathcal{SCQ}$ & \small $\mathcal{TF}$ & \small $\mathcal{OQ}$ & \small $\mathcal{SCQ}$ & \small $\mathcal{TF}$ & \small $\mathcal{OQ}$ & \small $\mathcal{SCQ}$ & \small $\mathcal{TF}$ & \small $\mathcal{OQ}$ \\
\midrule

Random & - & 26.25& 50.62& -& 23.28& 49.37& -& 29.44& 48.06& -& 26.67& 46.67& -& 26.67& 50.26& -\\
\midrule

\rowcolor{graybg} \multicolumn{17}{c}{\textit{Closed-source Models}} \\
\midrule
Gemini 3 Pro   & Direct & 60.39 & 50.00 & 33.23 & 61.70 & 64.56 & 35.14 & \bestD{80.95} & 74.27 & 60.34 & \bestD{95.56} & 76.67 & 67.59 & 53.33 & 67.69 & 48.02\\
Gemini 3 Flash & Direct & \bestD{61.44} & 59.38 & 33.87 & 61.81 & 70.89 & \bestD{37.84} & 79.22 & 69.90 & 60.51 & 91.11 & 70.00 & 70.34 & 60.00 & 69.74 & 49.04\\
GPT-5.2 Instant  & Direct & 55.87 & 56.25 & 36.45 & 57.80 & 62.03 & 32.79 & 76.19 & 69.90 & \bestD{70.77} & 73.33 & 80.00 & 67.93 & \bestD{66.67} & 73.85 & 38.76 \\
GPT-5 mini       & Direct & 57.72 & 62.50 & \bestD{44.52} & \bestD{62.61} & \bestD{78.48} & 35.50 & 67.10 & 74.76 & 55.56 & 71.11 & 70.00 & 63.45 & 33.33 & 55.90 & 26.10\\
\midrule

\rowcolor{graybg} \multicolumn{17}{c}{\textit{Open-source Models (Direct vs. LoRA)}} \\
\midrule

\multirow{2}{*}{LLaVA-1.5-7B} & Direct & 44.83 & 56.25 & 40.65 & 48.74 & 45.57 & 30.09 & 60.61 & 62.14 & 45.30 & 26.67 & 80.00 & 22.07 & 13.33 & \bestD{75.38} & 27.01\\
 & LoRA & 60.16 & 53.12 & \bestL{68.06} & 71.79 & 40.51 & 48.83 & 67.53 & 63.11 & 49.74 & 44.44 & 90.00 & 32.76 & 73.33 & \bestL{98.97} & 80.11\\
\cmidrule{1-17}

\multirow{2}{*}{LLaVA-NeXT-7B} & Direct & 40.77 & 56.25 & 35.81 & 49.20 & 54.43 & 28.83 & 58.87 & 61.65 & 48.38 & 44.44 & \bestD{83.33} & 28.62 & 46.67 & 64.10 & 34.12\\
 & LoRA & 63.41 & 78.12 & 62.58 & 79.01 & 77.22 & \bestL{53.15} & 79.65 & 83.50 & 55.73 & 51.11 & 93.33 & 41.72 & \bestL{86.67} & 95.38 & 79.10\\
\cmidrule{1-17}

\multirow{2}{*}{GLM-4.6V-Flash} & Direct & 53.08 & \bestD{71.88} & 41.29 & 54.70 & 70.89 & 33.51 & 67.10 & 68.93 & 61.71 & 75.56 & 60.00 & 64.48 & 46.67 & 46.67 & 28.93 \\
 & LoRA & 67.71 & \bestL{81.25} & 59.03 & 70.87 & 81.01 & 47.93 & 77.92 & 77.18 & 67.52 & 73.33 & 83.33 & 68.62 & 80.00 & 76.41 & 55.48\\
\cmidrule{1-17}

\multirow{2}{*}{InternVL3.5-2B} & Direct & 49.83 & 62.50 & 31.29 & 60.89 & 67.09 & 17.84 & 66.23 & 68.45 & 42.05 & 64.44 & 76.67 & 55.17 & 26.67 & 43.08 & 19.77\\
 & LoRA & 60.98 & \bestL{81.25} & 52.58 & 74.89 & 79.75 & 41.08 & 78.79 & 82.04 & 53.16 & 62.22 & 90.00 & 61.03 & 80.00 & 92.82 & 75.71\\
\cmidrule{1-17}

\multirow{2}{*}{InternVL3.5-8B} & Direct & 52.85 & 56.25 & 33.55 & 58.72 & 64.56 & 20.90 & 71.43 & 65.05 & 44.79 & 71.11 & 73.33 & 62.07 & 46.67 & 54.36 & 24.86\\
 & LoRA & 64.69 & 78.12 & 58.39 & 78.56 & 79.75 & 46.13 & 83.55 & 84.47 & 61.54 & 66.67 & \bestL{96.67} & 61.72 & 80.00 & 94.36 & 80.45\\
\cmidrule{1-17}

\multirow{2}{*}{InternVL3.5-14B} & Direct & 47.62 & 65.62 & 31.61 & 58.83 & 64.56 & 24.14 & 71.00 & 69.90 & 51.62 & 71.11 & \bestD{83.33} & 58.28 & 46.67 & 49.74 & 22.94\\
 & LoRA & 67.71 & 78.12 & 50.97 & 78.33 & 78.48 & 47.03 & 84.42 & 86.41 & 68.72 & \bestL{77.78} & 86.67 & 61.03 & \bestL{86.67} & 89.23 & 80.90\\
\cmidrule{1-17}

\multirow{2}{*}{Qwen3-VL-8B} & Direct & 54.36 & 62.50 & 37.74 & 57.68 & 62.03 & 32.97 & 73.16 & \bestD{76.70} & 62.05 & 73.33 & 76.67 & 71.38 & 53.33 & 58.97 & 45.20\\
 & LoRA & 73.17 & 78.12 & 63.55 & 81.31 & 81.01 & 51.17 & 80.52 & \bestL{89.32} & 68.03 & \bestL{77.78} & 93.33 & 70.34 & 73.33 & 91.79 & 77.97\\
\cmidrule{1-17}

\multirow{2}{*}{Qwen3-VL-32B} & Direct & 57.61 & 65.62 & 35.81 & 59.98 & 67.09 & 30.09 & 74.89 & 68.93 & 62.56 & 80.00 & 80.00 & \bestD{78.97} & 60.00 & 65.13 & \bestD{52.43}\\
 & LoRA & \bestL{73.64} & 75.00 & 64.52 & \bestL{81.65} & \bestL{88.61} & 50.45 & 82.68 & 88.83 & \bestL{72.31} & \bestL{77.78} & 86.67 & \bestL{74.14} & 80.00 & 85.13 & \bestL{81.24}\\

\bottomrule
\end{tabular}%
}
\end{table*}

\subsection{Experimental Setup}
\label{sec:exp_setup}

We conduct a comprehensive evaluation across a wide spectrum of MLLMs, spanning both proprietary and open-source architectures. For proprietary models, we test the latest iterations including Gemini 3 (Pro/Flash) \cite{geminiteam2025geminifamilyhighlycapable} and the GPT-5 series (5.2-Instant/5-Mini) \cite{gpt5_system_card}. For open-source models, we select representative baselines with varying scales: InternVL3.5 (2/8/14B) \cite{wang2025internvl3_5}, Qwen3-VL (8/32B) \cite{bai2025qwen3vltechnicalreport}, LLaVA v1.5 \cite{liu2023improvedllava}, LLaVA-NeXT \cite{liu2024llavanext}, and GLM-4.6v-Flash \cite{vteam2025glm45vglm41vthinkingversatilemultimodal}. To establish performance bounds, we incorporate a random baseline for choice-based tasks and report human performance evaluated on 1,000 samples (balanced between simulation and real-world data) by three volunteers. 
The evaluation operates under two settings: (1) \textbf{Zero-shot Inference}, where models directly predict answers from visual-textual inputs; and (2) \textbf{Instruction Tuning}, where we apply LoRA-based \citep{hu2022lora} parameter-efficient fine-tuning. Crucially, our training set consists exclusively of simulation data to assess sim-to-real generalization. Implementation details are provided in Appendix~\ref{sec:app_ex}.

\subsection{Evaluation Metrics}
\label{sec:metrics}

\texttt{EgoPoint-Bench} comprises three task types: True/False (TF), Single Choice Questions (SCQ), and Open-ended Questions (OQ). Following established protocols \citep{fu2025video,li2024mvbench}, we adopt exact matches for the  TF and SCQ tasks. For the OQ task, evaluating open-ended responses remains challenging; therefore, we employ an LLM-as-a-Judge approach \cite{zheng2023judging}. Specifically, GPT-4o \cite{gpt4o} scores the model predictions against ground-truth answers on a scale of 0 to 1 (with an increment of 0.2). Further details can be found in Appendix \ref{sec:app_eval}.

\subsection{Main Results}
\label{sec:main_results}

Table~\ref{tab:main_results_final} presents the performance of proprietary and open-source models across simulation and real-world test sets. We reported three key observations:

\noindent \textbf{Off-the-shelf VLMs struggle with fine-grained egocentric deictic understanding.} In the Direct inference setting, even the most advanced proprietary models (e.g., Gemini 3 Pro, GPT-5 mini) and open-source models fail to achieve satisfactory performance, hovering around 60\% accuracy overall. A significant gap remains compared to human performance (95.90\%), particularly in tasks requiring precise spatial geometric reasoning (AR and BP metrics). This underscores that general-purpose pre-training is insufficient for comprehending complex ``finger-pointing'' semantics in egocentric views.

\noindent \textbf{Simulation-based tuning yields significant gains.} Fine-tuning with our generated simulation data via LoRA brings substantial improvements across all open-source models. As shown in the ``Gain'' column, we observe a consistent performance boost ranging from +15.27\% to +25.41\%. Notably, LLaVA-Next-7B achieves a remarkable 25.41\% improvement, demonstrating that the visual-semantic alignment provided by our synthetic data effectively unlocks the models' potential for pointing-oriented VQA tasks.

\noindent \textbf{Effective Sim-to-Real generalization.} Crucially, the models trained on simulation data generalize exceptionally well to the Real-world testset. For instance, Qwen3-VL-8B improves its real-world mean accuracy from 58.14\% to 71.96\% after tuning on simulation data. This suggests that the geometric and semantic features of finger-pointing learned from our high-fidelity simulation environment are robust and transferrable, validating the efficacy of our data generation pipeline for real-world applications.

\subsection{Detailed Analysis}
\label{sec:detailed_analysis}

\noindent\textbf{Analysis Across Different Question Types.} Table \ref{tab:detailed_results} dissects model performance across three answer formats ($\mathcal{SCQ}, \mathcal{TF}, \mathcal{OQ}$), revealing three critical insights: (1) \textbf{Generative bottleneck.} Direct models exhibit a sharp performance drop in Open-Ended questions ($\mathcal{OQ}$) compared to discriminative formats ($\mathcal{SCQ}, \mathcal{TF}$), indicating that while pre-trained models can distinctively \textit{recognize} correct references, they struggle to actively \textit{formulate} precise spatial descriptions without specific tuning. (2) \textbf{Geometric alignment in Adversarial Relations.} The AR dimension, which requires distinguishing targets from spatial distractors, sees the most dramatic gains from LoRA (e.g., Llava-1.5-7B AR-$\mathcal{OQ}$ jumps from 27.01\% to 80.11\%). 
This suggests that our dataset helps models better capture pointing-related spatial cues that are underrepresented in general pretraining.
(3) \textbf{Spatial-semantic saturation.} 
While models show a high baseline and limited room for improvement in text-heavy tasks (OCR), they experience dramatic gains in spatial reasoning tasks (BP, SC, AR). This contrast highlights that our approach primarily enhances fine-grained spatial capabilities rather than basic visual recognition.

\begin{table}[t]
\centering
\definecolor{graybg}{rgb}{0.94, 0.94, 0.94}
\setlength{\aboverulesep}{0pt}
\setlength{\belowrulesep}{0pt}
\renewcommand{\arraystretch}{1.3}

\caption{Performance evaluation of representative MLLMs on Sim and Real test sets across three deixis levels (L1-L3). The best results are highlighted in \textbf{bold}.}
\label{tab:results_deixis}

\resizebox{\columnwidth}{!}{%
\begin{tabular}{l|c|ccc|ccc}
\toprule
\multirow{2}{*}{\textbf{Model}} & \multirow{2}{*}{\textbf{Method}} & \multicolumn{3}{c|}{\textbf{Sim}} & \multicolumn{3}{c}{\textbf{Real}} \\
 & & \textbf{L1} & \textbf{L2} & \textbf{L3} & \textbf{L1} & \textbf{L2} & \textbf{L3} \\
\midrule

Gemini 3 Pro   & Direct & 51.03& 59.00& 59.53& \textbf{72.57}& 65.20& 72.76\\
GPT-5 mini       & Direct & 58.22& 59.82& 56.02& 59.32& 54.40& 64.38\\
\midrule

\multirow{2}{*}{InternVL3.5-2B} & Direct & 52.51& 53.83& 49.96& 56.25& 48.20& 50.73\\
 & LoRA & 74.71& 74.60& 76.47& 59.03& 61.40& 67.50\\
\cmidrule{1-8}

\multirow{2}{*}{Llava-1.5-7B} & Direct & 48.11& 51.47& 47.98& 42.27& 52.60& 54.48 \\
 & LoRA & 72.01& 75.60& 72.83& 51.21& 60.80& 58.80\\
\cmidrule{1-8}

\multirow{2}{*}{Qwen3-VL-32B} & Direct & 50.52& 62.72& 62.28& 64.31& 62.40& 64.79\\
 & LoRA & \textbf{83.77}& \textbf{81.63}& \textbf{81.20}& 69.59& \textbf{71.80}& \textbf{75.83}\\

\bottomrule
\end{tabular}%
}
\end{table}

\paragraph{Impact of different deixis levels.} 
An analysis of model performance across the three deixis levels (L1, L2, L3) in Table \ref{tab:results_deixis} reveals a distinct progression from weak zero-shot alignment to robust, fine-tuned generalization. In the zero-shot Direct setting, off-the-shelf MLLMs demonstrate a weak alignment between explicit linguistic instructions and geometric visual cues; for instance, models like Gemini 3 Pro and Qwen3-VL-32B often perform better on vague locatives (L2) or implicit pronouns (L3) than on explicit action descriptions (L1). However, following LoRA fine-tuning on our synthetic data, this performance gap narrows significantly. In the Sim domain, fine-tuned models achieve highly balanced and elevated scores across all deixis levels, demonstrating that spatially-aware supervision successfully teaches the precise alignment of explicit language with fine-grained pointing kinematics. 
Crucially, this capability demonstrates robust Sim-to-Real generalization: on the unconstrained Real-world dataset, our fine-tuned models exhibit substantial improvements across all three levels (L1-L3). 
These results support the view that spatial reasoning learned from simulation can transfer to real-world settings, reducing reliance on visual saliency or scene priors.

\subsection{Error Types}
\begin{figure}[t]
  \includegraphics[width=\columnwidth]{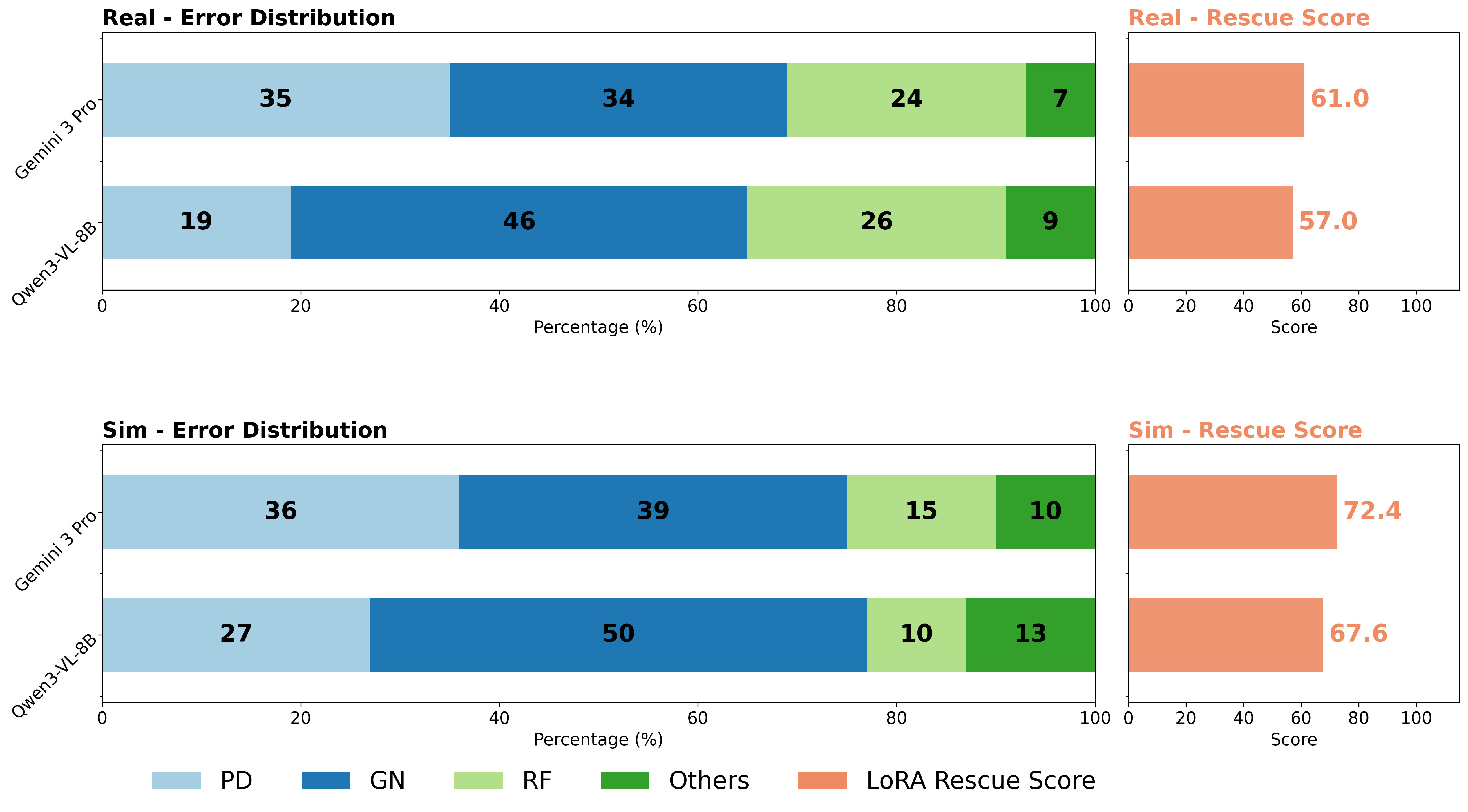}
  \caption{Distribution of error types and rescue scores.}
  \label{fig:error_analysis}
\end{figure}

To probe the limitations of current VLMs in finger-pointing VQA, we conducted a manual analysis on 400 error cases generated by Qwen3-VL-8B and Gemini 3 Pro (balanced between simulated and real-world data). We classified errors into three primary categories: (1) \textbf{Proximal Distraction (PD)}, where the model fails to follow the pointing ray and instead grounds the answer to a distractor immediately adjacent to the finger; (2) \textbf{Gesture Neglect (GN)}, where the model ignores the gesture entirely, attending to visually salient or distant objects; and (3) \textbf{Reasoning Failure (RF)}, where the target is correctly localized, but the model fails in downstream reasoning. Fig. \ref{fig:error_analysis} (Left) illustrates the error distribution, revealing that PD and GN are the most prevalent failure modes. Fig. \ref{fig:error_analysis} (Right) demonstrates the efficacy of our approach by reporting the ``Rescue Score''—defined as the percentage of these specific failure cases successfully corrected by our LoRA-finetuned Qwen3-VL-8B. Our method achieves Rescue Scores ranging from 57.0\% to 72.4\% across datasets, confirming its capability to effectively recover from the spatial ambiguity and gesture perception issues inherent in the baselines.
\begin{figure}[ht]
  \includegraphics[width=\columnwidth]{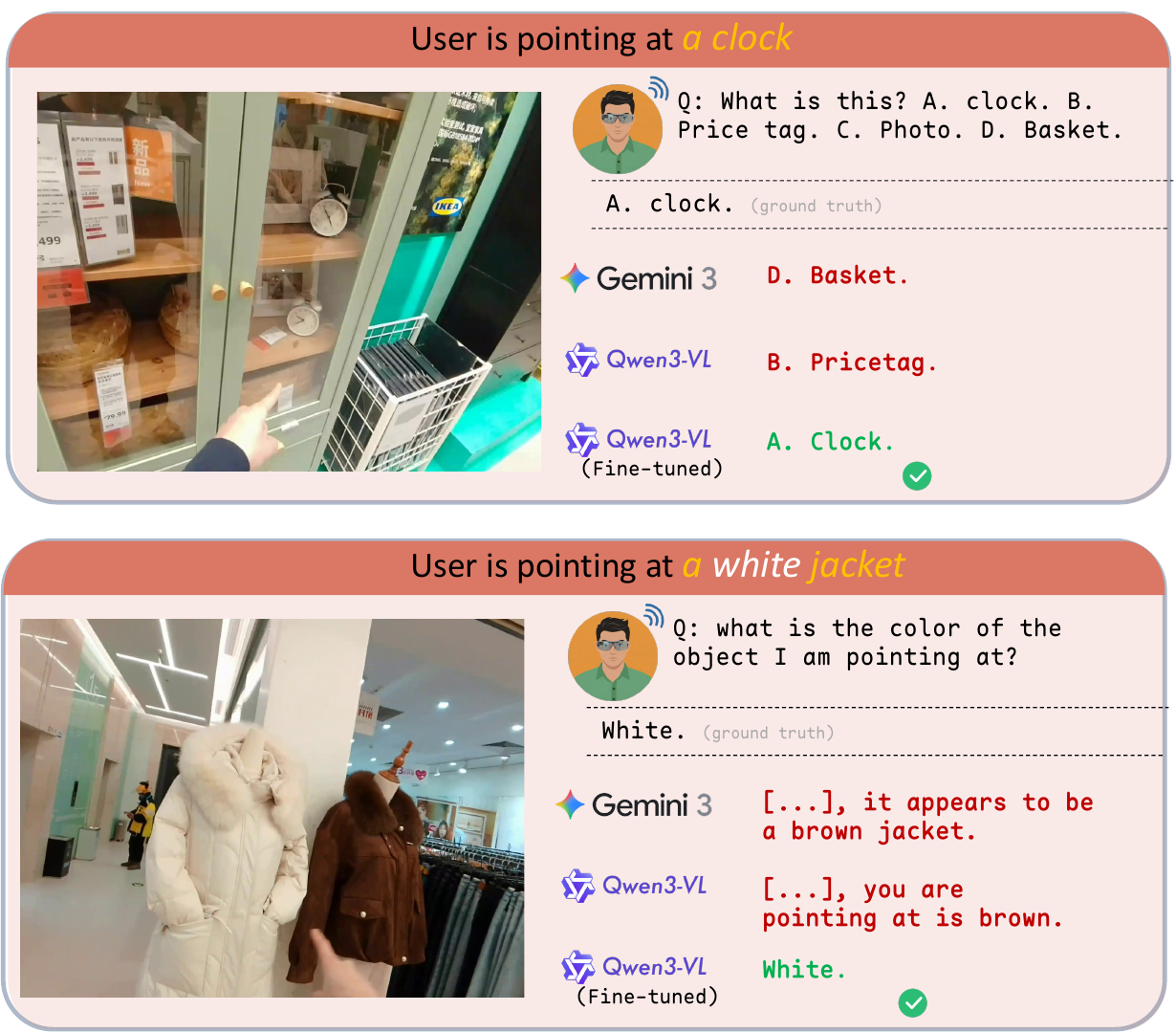}
  \caption{Comparison of model performance on real-world pointing tasks.}
  \label{fig:error_lora}
\end{figure}

Fig. \ref{fig:error_lora} presents two examples of random inquiries conducted in real-world environments. In the first example, both Gemini 3 Pro and Qwen3-VL-8B provide incorrect and inconsistent answers, highlighting their tendency to make arbitrary guesses in the background when the reference is unclear. In the second example, featuring a white and a brown jacket, the user points toward the white one; however, due to perspective effects, the finger region appears closer to the brown jacket in the image. Consequently, both base models consistently fail this task. In contrast, our Qwen3-VL-8B model, fine-tuned with LoRA on simulation data, is able to answer both questions with complete accuracy. More examples are provided in Appendix~\ref{sec:error_ana}.

\section{Conclusion}
We introduced \texttt{EgoPoint-Bench} to evaluate and enhance MLLMs' understanding of egocentric finger-pointing gestures. Our evaluation reveals that while existing MLLMs struggle with this task, fine-tuning on high-quality synthetic data mitigates referential hallucinations, enabling robust real-world generalization. This work paves a scalable path toward precise egocentric AI assistants.

\section*{Limitations}

While \texttt{EgoPoint-Bench} provides a benchmark for evaluating current egocentric multimodal finger-pointing understanding, it possesses two primary limitations: 1) Although fine-tuning with automatically synthesized simulation data has proven effective on real-world datasets, we observed that the performance gain on real-world data is smaller than that on simulated data. This suggests that real-world user pointing behaviors, along with environmental complexities such as arm backgrounds, are significantly more intricate and challenging than those in simulation. Simulated data struggles to sufficiently cover the behavioral characteristics of the real world. 2) To facilitate easier evaluation, current dataset questions and answers are relatively brief, which diverges from the complex, multi-turn dialogue patterns found in real-world interactions. We focus first on whether MLLMs can explicitly understand the fundamental meaning of ``pointing,'' as our experimental results indicate that even this poses a significant challenge for current models. Mastering these basic comprehension skills is a vital prerequisite before addressing more difficult and complex multi-turn interaction tasks.

\section*{Ethical Statement}
University ethics review board approves human-subjects research and they approved this project. In our real-world data collection environment, we have anonymized all human faces and any identifying information within the images by applying a blurring treatment. This ensures that no privacy leaks occur and that the dataset contains no harmful content. All datasets used in this work, including HM3D, AI2-THOR, ReplicaCAD, and HSSD, are properly cited and used strictly for non-commercial academic research purposes.

\section*{Acknowledgments}
This work was supported in part by the National Natural Science Foundation of China under Grant 62376132.

\bibliography{custom}

\newpage
\appendix

\section{Experimental Setup}

\label{sec:app_ex}
\subsection{Model Configurations}
Regarding the configurations of the mainstream MLLMs we evaluated: specifically, for the \texttt{Qwen3-VL} and \texttt{InternVL3.5} series, we utilized their \texttt{Instruct} variants. Furthermore, for all open-source models, we set \texttt{Do Sample=False} during inference; and for all closed-source models, we set \texttt{Temperature=0.0} and \texttt{Top-P=1}. This implies that we employed deterministic decoding strategies (i.e., greedy search) to eliminate randomness during generation, thereby ensuring the reproducibility of the evaluation results and fairness in comparisons across different models.

\subsection{Additional Implementation Details}
To systematically evaluate the performance of Multi-modal Large Language Models (MLLMs) on \texttt{EgoPoint-Bench}, we utilized the official open-source implementations of each model. All evaluation experiments and instruction tuning processes were conducted on NVIDIA A100 GPUs. Our evaluation framework is built upon the Hugging Face Transformers library\footnote{\url{https://huggingface.co/docs/transformers}} and leverages the LLaMA-Factory framework \cite{zheng2024llamafactory} for efficient fine-tuning.

To ensure fair comparison and reproducibility, we standardized training configurations across all models using LoRA ($r=8$) applied to all linear layers. We utilized a global batch size of 64 (per-device batch size 8 with 8 accumulation steps), enabled bfloat16 precision, and trained for 3 epochs with a learning rate of $1 \times 10^{-4}$ using a Cosine learning rate scheduler.

\subsection{Curated Prompt Templates}
The text data utilized for both zero-shot inference and LoRA fine-tuning remains consistent across all models, formatted as follows:

\begin{tcolorbox}[
    breakable,
    enhanced,
    colback=bgGray,      
    colframe=gray!30,    
    boxrule=1pt,         
    arc=4mm,             
    title=\textbf{Prompt Templates}, 
    coltitle=black,      
    colbacktitle=gray!10,
    fontupper=\ttfamily\small, 
    left=6mm, right=6mm, top=4mm, bottom=4mm 
]

    \promptkey{Single Choice} \\
    \promptuser: \promptvar{Question} \textbackslash n \promptvar{Options} \textbackslash n Answer directly using the letters of the options given.
    
    \vspace{0.4cm} 
    
    \promptkey{True/False} \\
    \promptuser: \promptvar{Question} \textbackslash n Answer directly with `True' or `False'
    
    \vspace{0.4cm} 
    
    \promptkey{Open Ended} \\
    \promptuser: \promptvar{Question} \textbackslash n Please output the answer directly.

\end{tcolorbox}

\subsection{Scoring Open-ended Question}
\label{sec:app_eval}
We use the following carefully crafted prompts to score each open-ended question:
\definecolor{judgeBlue}{RGB}{0, 51, 102}
\definecolor{criteriaGreen}{RGB}{0, 102, 51}
\definecolor{varRed}{RGB}{153, 0, 0}
\definecolor{jsonOrange}{RGB}{204, 102, 0}
\definecolor{bgGray}{gray}{0.98}

\begin{tcolorbox}[
    breakable,
    enhanced,
    colback=bgGray,      
    colframe=gray!30,    
    boxrule=1pt,         
    arc=4mm,             
    title=\textbf{Evaluation Prompt Template}, 
    coltitle=black,      
    colbacktitle=gray!10,
    fontupper=\ttfamily\small, 
    left=6mm, right=6mm, top=4mm, bottom=4mm
]

    {\color{judgeBlue}\textbf{Role:}} You are a helpful assistant evaluation judge. \\
    Please evaluate the candidate answer against the reference answer based on the question. \\
    Assign a score from 0 to 5.

    \vspace{0.3cm}

    {\color{criteriaGreen}\textbf{Scoring Criteria:}} \\
    \textbf{0}: Completely incorrect or irrelevant. \\
    \textbf{1}: Contains some keywords but fails to answer the question logic. \\
    \textbf{2}: Partially correct but misses key constraints. \\
    \textbf{3}: Mostly correct, but contains minor hallucinations or ambiguity. \\
    \textbf{4}: Correct meaning, but phrased awkwardly or includes unnecessary fluff. \\
    \textbf{5}: Perfect match in meaning and accuracy.

    \vspace{0.3cm}

    \textbf{Input:} \\
    Question: {\color{varRed}\{question\}} \\
    Reference Answer: {\color{varRed}\{answer\}} \\
    Candidate Answer: {\color{varRed}\{model\_output\}}

    \vspace{0.3cm}

    \textbf{Output Format:} \\
    You MUST return a valid JSON object strictly adhering to the following structure: \\
    {\color{jsonOrange}
    \{ \\
    \hspace*{1em} "score": <integer\_0\_to\_5>, \\
    \hspace*{1em} "reason": "<short\_explanation\_string>" \\
    \}
    }

\end{tcolorbox}

\section{Additional Analysis}
\subsection{Detailed Dataset Statistics}

\begin{figure*}[t]
\centering
  \includegraphics[width=\linewidth]{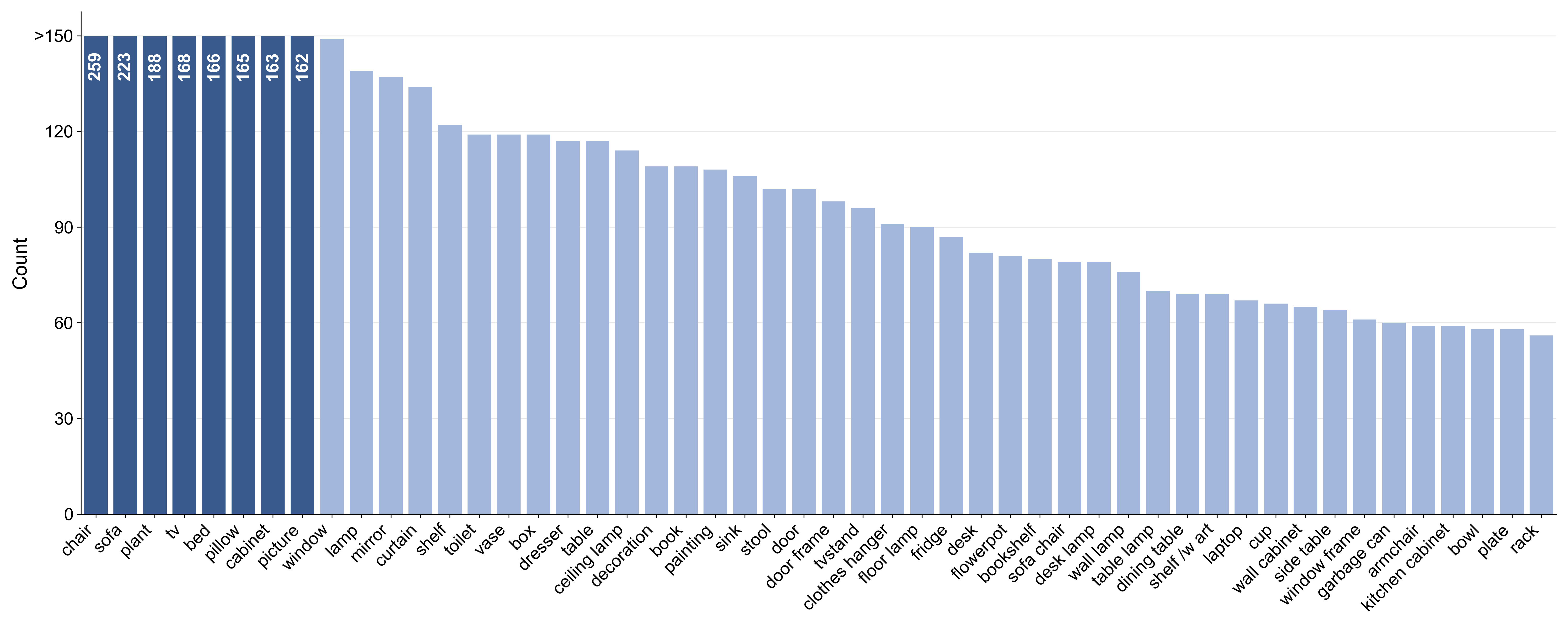}
  \caption{Frequency of  top-50 object categories in simulation data.} 
\label{fig:sim-data-top50}
\end{figure*}

Fig. \ref{fig:sim-data-top50} illustrates the top 50 most frequent object categories in the simulation dataset. These categories primarily encompass complex indoor scenes, where high spatial coupling and environmental complexity pose significant challenges for model understanding. Consequently, the dataset demonstrates high sample diversity and task difficulty.

\begin{figure}[ht]
  \includegraphics[width=\columnwidth]{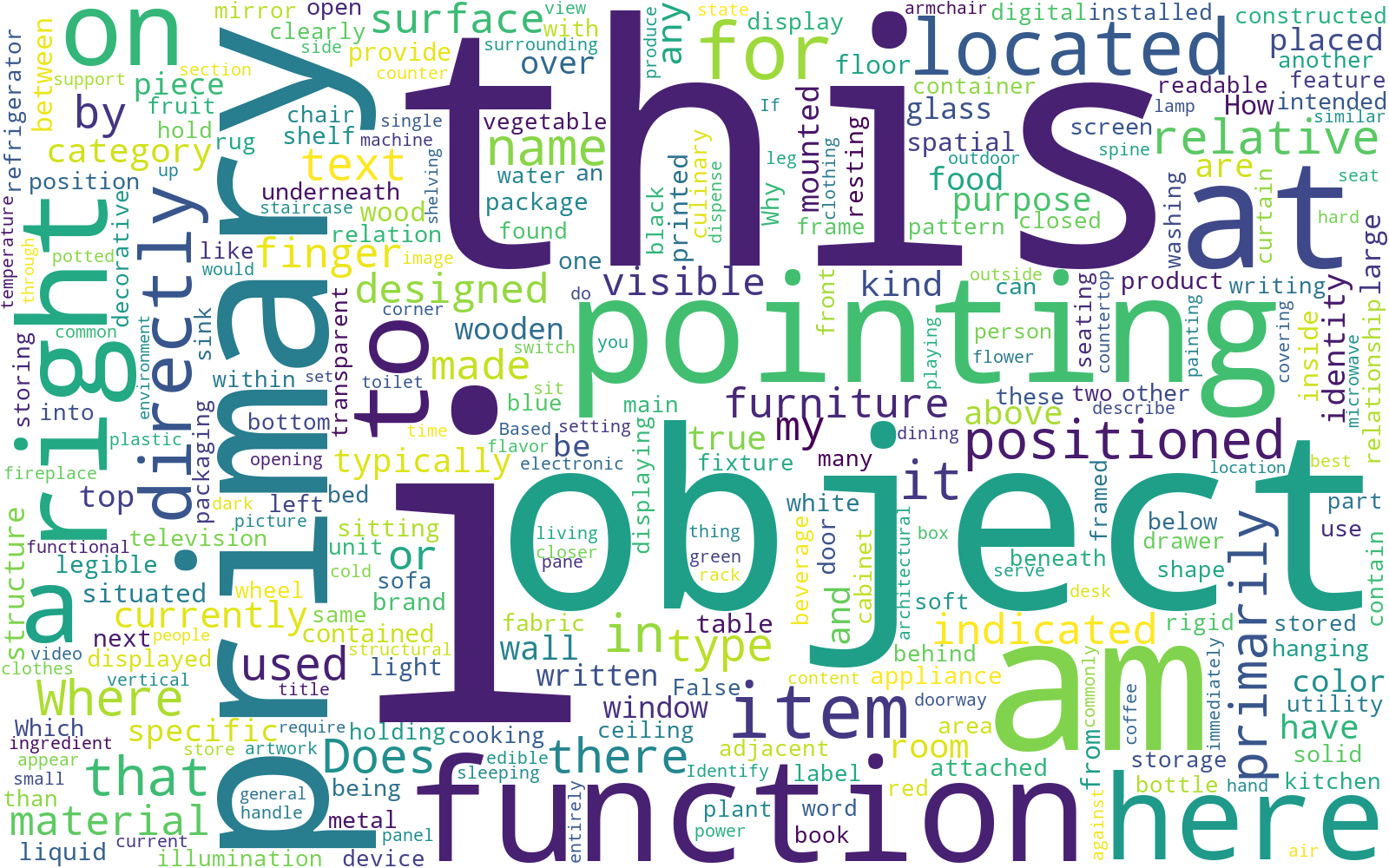}
  \caption{Word cloud of questions in \texttt{EgoPoint-Bench}.}
  \label{fig:wordcloud}
\end{figure}

Fig. \ref{fig:wordcloud} illustrates the word cloud of all questions within \texttt{EgoPoint-Bench}. The results reveal a prevalence of deictic expressions (e.g., this, pointing at, here, that), indicating a strong emphasis on both explicit pointing and ambiguous reference. This distribution aligns perfectly with the core design philosophy of \texttt{EgoPoint-Bench}: to evaluate the model's capability in referential understanding during egocentric multimodal interactions.

Table \ref{tab:dataset_stats} provides a detailed breakdown of the data sources across the training, validation, and testing sets. Extensive samples were drawn from HM3D due to its high-fidelity rendering of real-world environments. Conversely, ReplicaCAD was sampled sparingly and utilized only for training and validation, given its limited variety of scenes and objects. Notably, real-world data was reserved exclusively for testing to evaluate zero-shot generalization. Furthermore, the average question length of 9.81 underscores the distinctive nature of deictic language in egocentric VQA tasks.

\begin{table}[htbp]
  \centering
  \caption{Dataset Statistics and Split Details}
  \label{tab:dataset_stats}
  \renewcommand{\arraystretch}{1.2} 
  \resizebox{\columnwidth}{!}{%
  \begin{tabular}{llccccc}
    \toprule
    \textbf{Source} & \textbf{Subset} & \textbf{Train} & \textbf{Val} & \textbf{Test} & \textbf{Total} & \textbf{Avg. Q Len.} \\
    \midrule
    \multirow{4}{*}{Sim} 
      & HM3D & 3227 & 365 & 718 & 4310 & 10.12 \\
      & HSSD & 1964 & 214 & 605 & 2783 & 8.68 \\
      & AI2-THOR & 1982 & 220 & 606 & 2808 & 10.22 \\
      & ReplicaCAD & 601 & 65 & - & 666 & 8.67 \\
    \midrule
    Real & -     & -  & -  & 1162  & 1162  & 11.02  \\
    \bottomrule
  \end{tabular}
  }

\end{table}

\begin{figure}[ht]
  \includegraphics[width=\columnwidth]{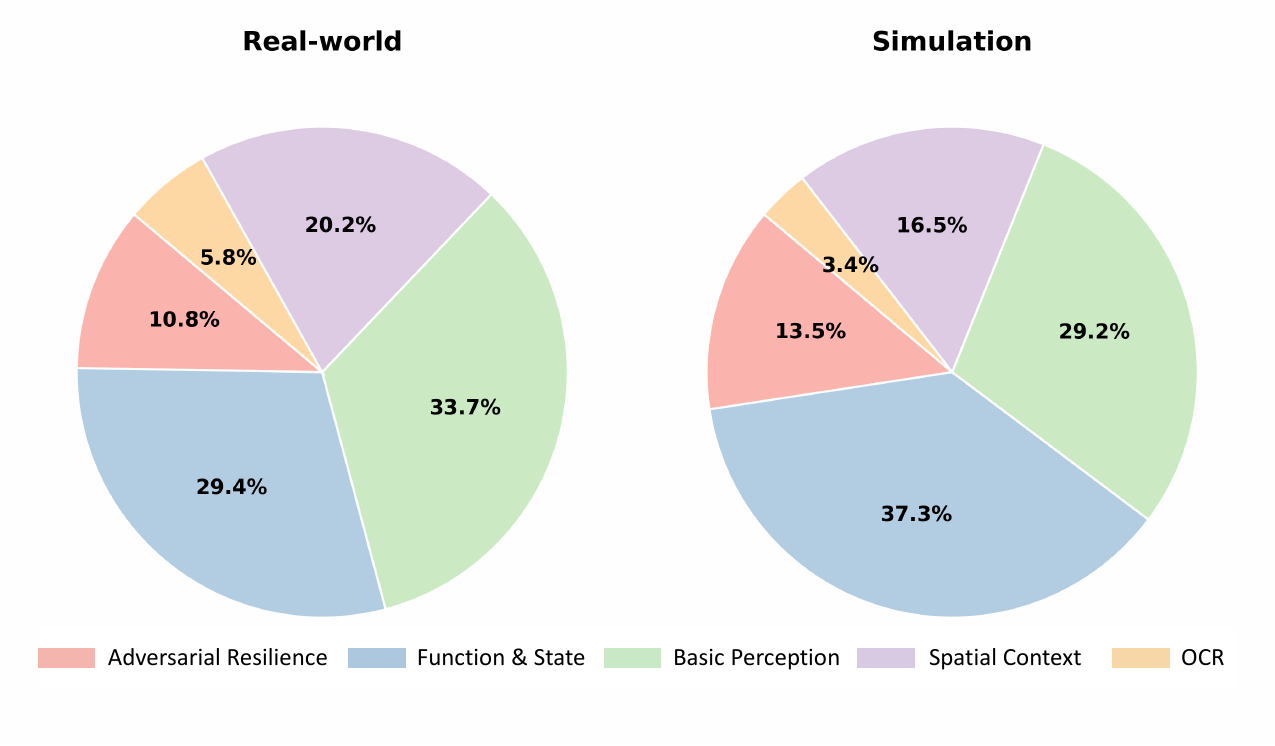}
  \caption{Distribution of 5 dimensions in \texttt{EgoPoint-Bench} testset.}
  \label{fig:dimension_distribution}
\end{figure}

\begin{figure}[ht]
  \includegraphics[width=\columnwidth]{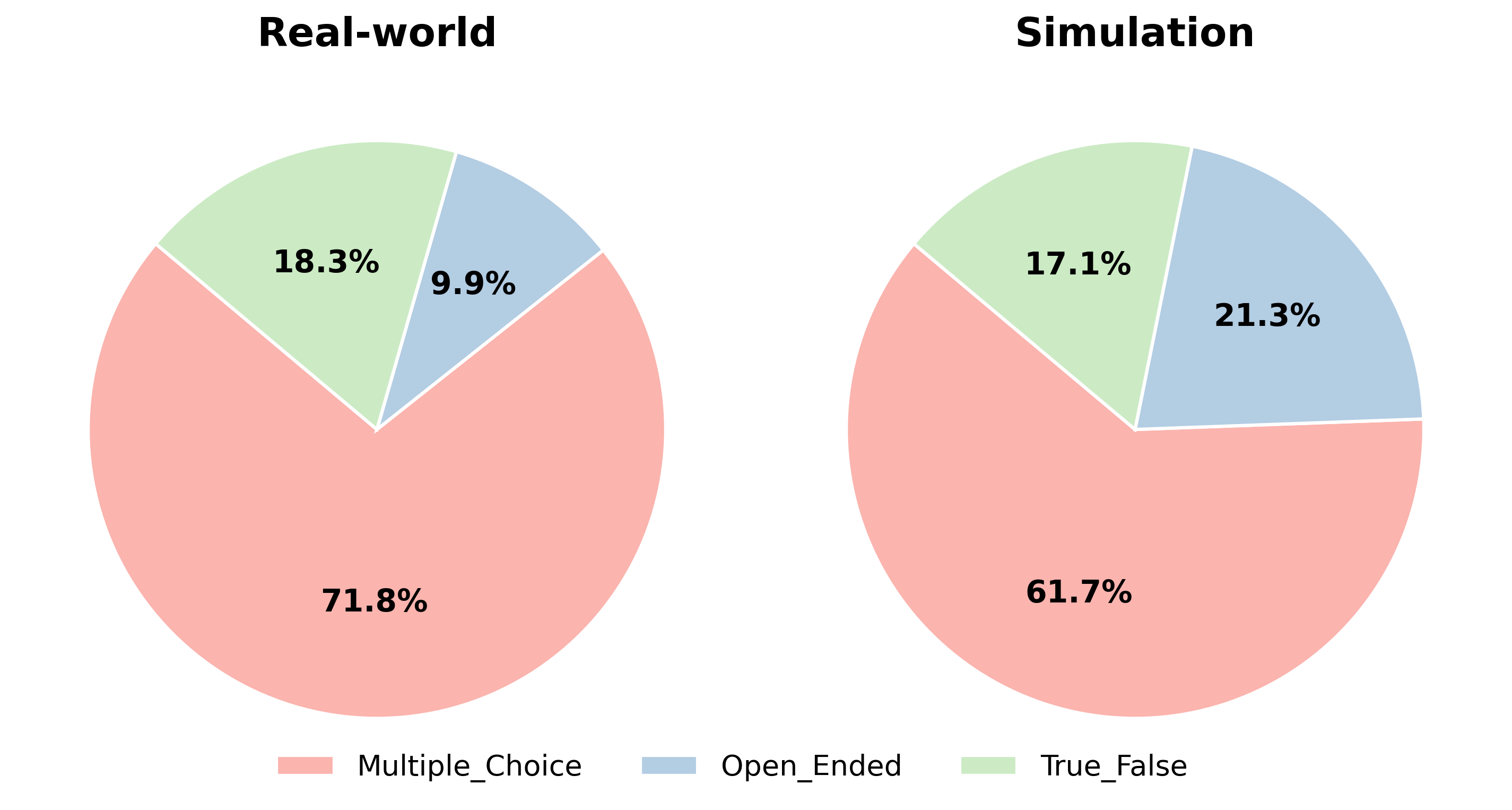}
  \caption{Distribution of 3 question types in \texttt{EgoPoint-Bench} testset.}
  \label{fig:type_distribution}
\end{figure}

Figs. \ref{fig:dimension_distribution} and \ref{fig:type_distribution} illustrate the distribution of question dimensions and types in the test set, respectively. The dataset primarily evaluates Basic Perception and Affordance, mirroring common queries in daily life regarding object attributes and functional utilities. To ensure objective benchmarking, the questions are predominantly binary and multiple-choice, while open-ended questions are included to better simulate real-world QA scenarios. 

\begin{figure}[ht]
  \includegraphics[width=\columnwidth]{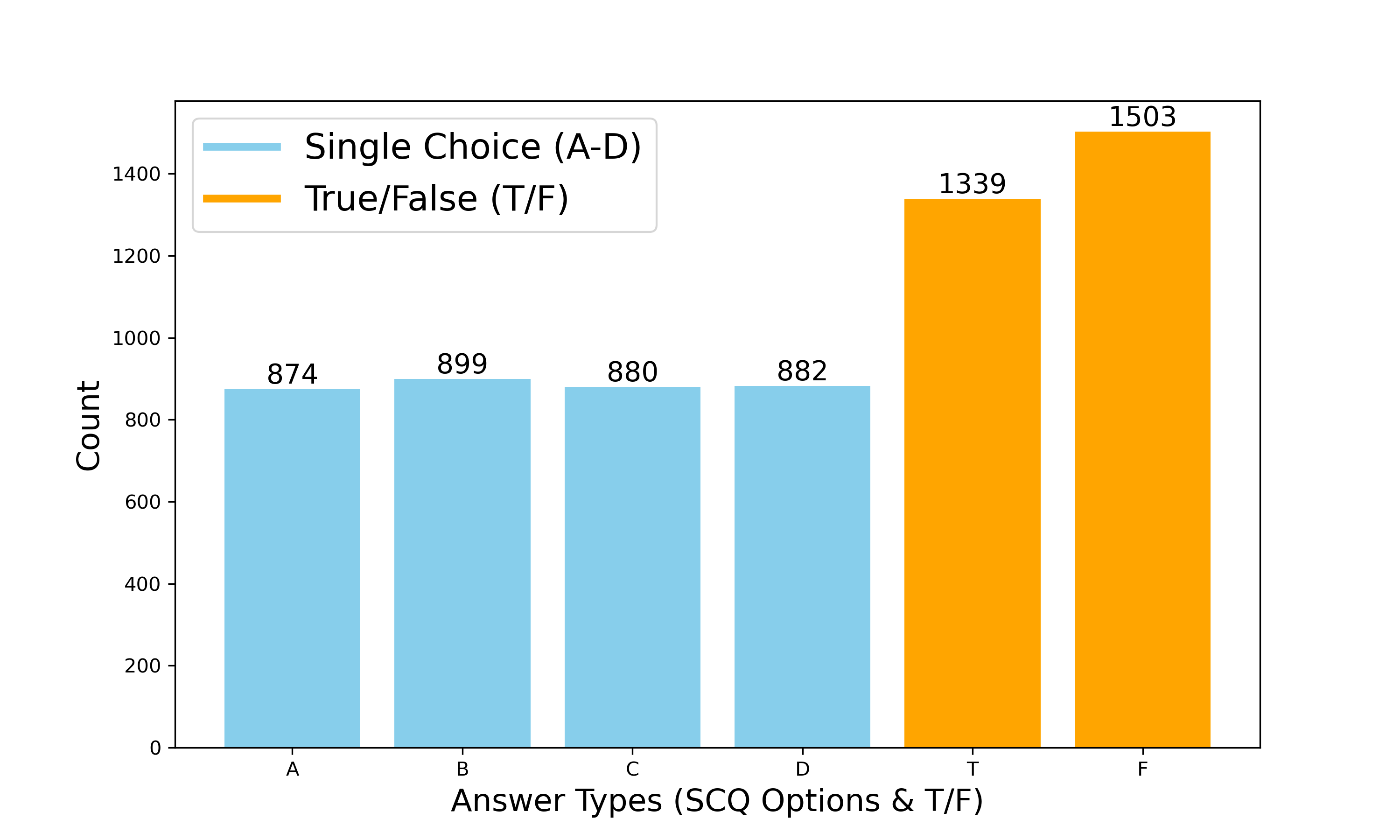}
  \caption{Option distribution of training set.}
  \label{fig:answer_stat}
\end{figure}

Furthermore, Fig. \ref{fig:answer_stat} shows a balanced distribution of question types in the training set, preventing the model from developing a preference bias toward specific answer labels.

\subsection{Error Analysis}
\label{sec:error_ana}

\begin{figure*}[ht]
  \includegraphics[width=\linewidth]{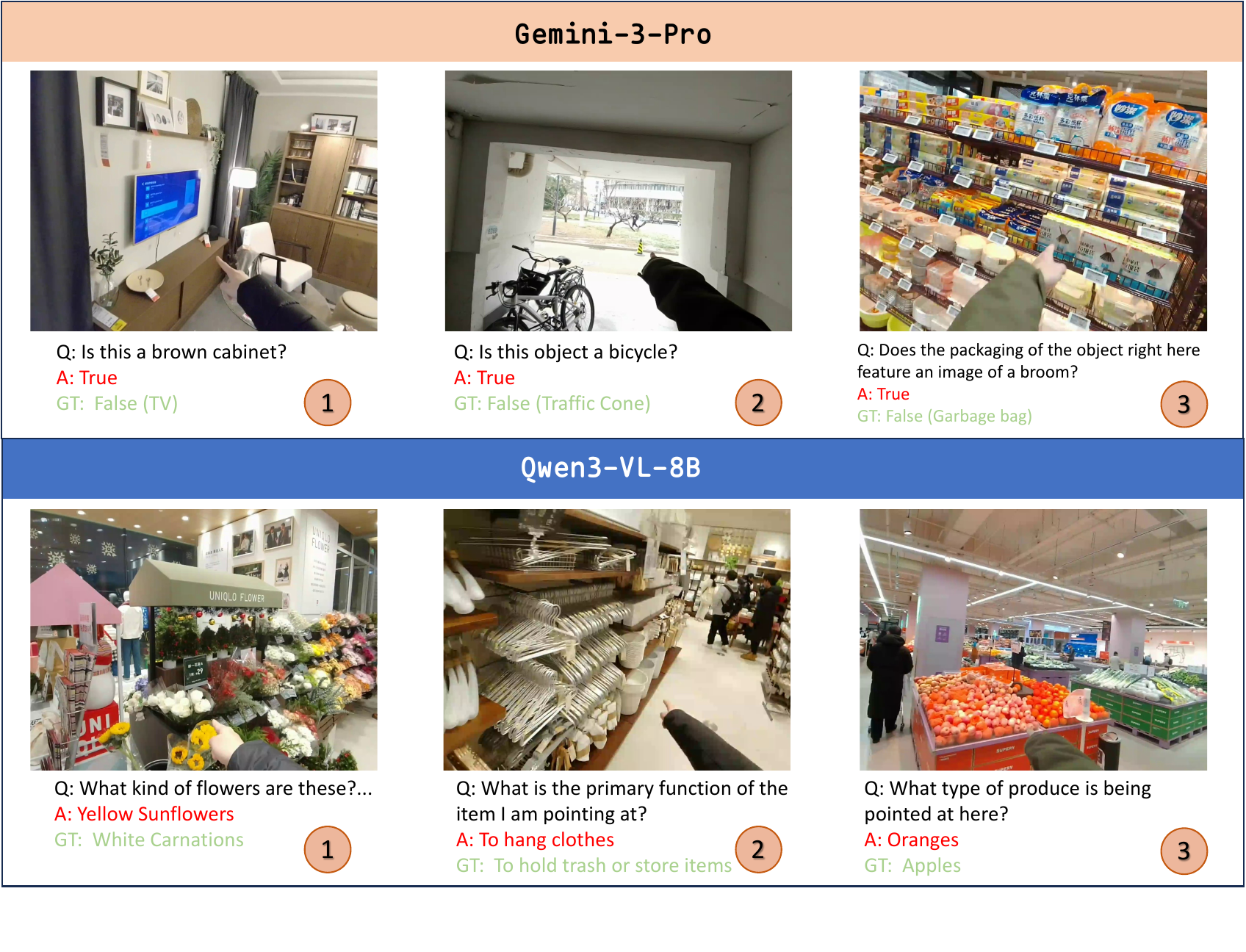}
  \caption{Error examples of three types in two methods from real-world data.}
  \label{fig:error_real}
\end{figure*}

\begin{figure*}[ht]
  \includegraphics[width=\linewidth]{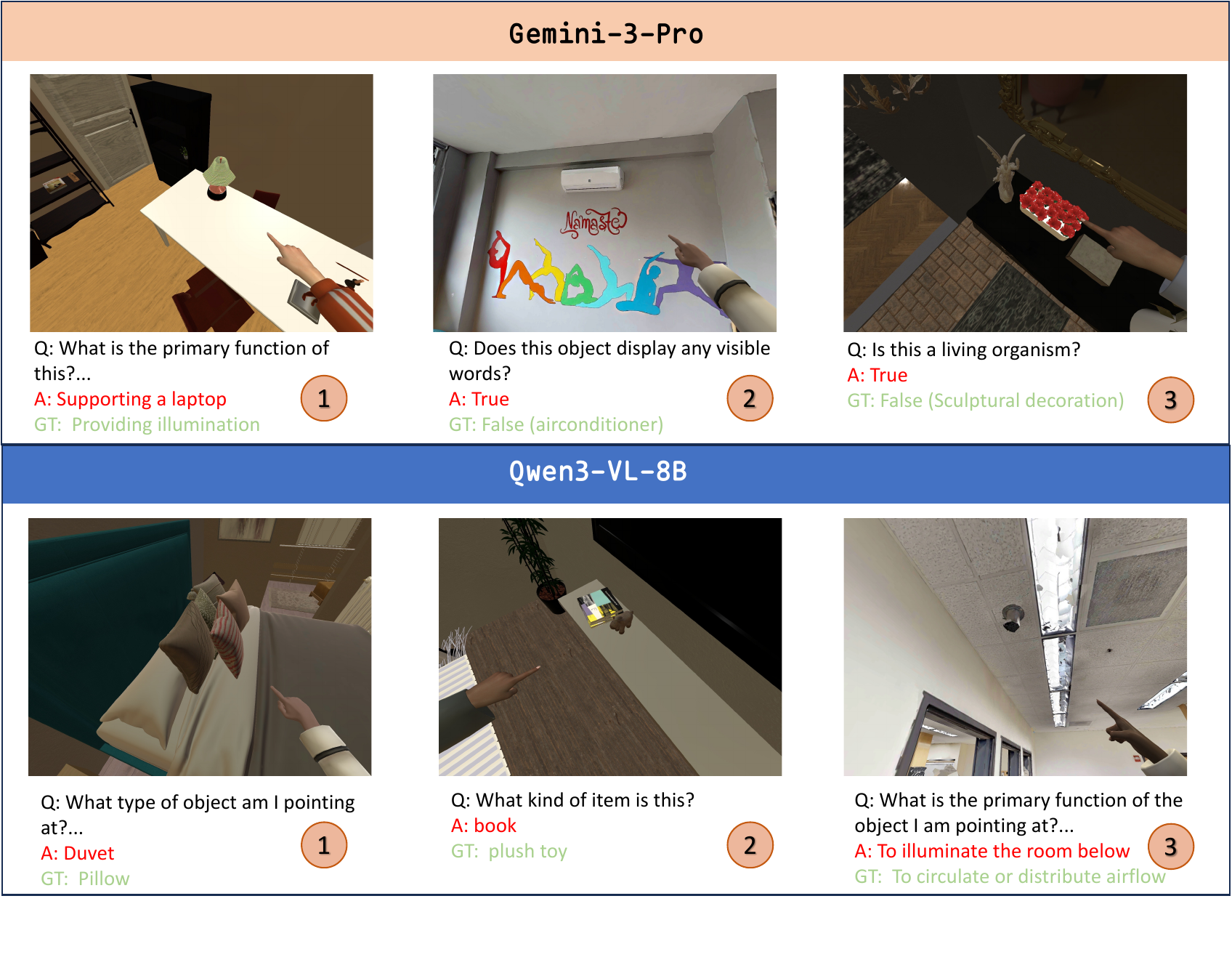}
  \caption{Error examples of three types in two methods from simulation data.}
  \label{fig:error_sim}
\end{figure*}

Figs. \ref{fig:error_real} and \ref{fig:error_sim} illustrate three representative error types made by Gemini 3 Pro and Qwen3-VL-8B on real-world and simulation datasets, respectively (where Q denotes the question, A the model’s response, and GT the ground-truth intent). The results indicate that these models are highly susceptible to interference from objects in close proximity to the hand or prominent objects in the background.

\subsection{Qualitative Success Cases Across Five Dimensions}
To further illustrate the capability of the fine-tuned model, we provide qualitative examples of Qwen3-VL-8B after LoRA fine-tuning on our synthetic data across all five evaluation dimensions, including \textit{Basic Perception}, \textit{Function \& State}, \textit{Spatial Context}, \textit{OCR}, and \textit{Adversarial Resilience}. For each dimension, we show three representative examples. These cases are intended to demonstrate the diversity of question types in EgoPoint-Bench and the effectiveness of the fine-tuned model in resolving pointing-based referential queries.

\begin{figure*}[t]
    \centering
    \includegraphics[width=\textwidth]{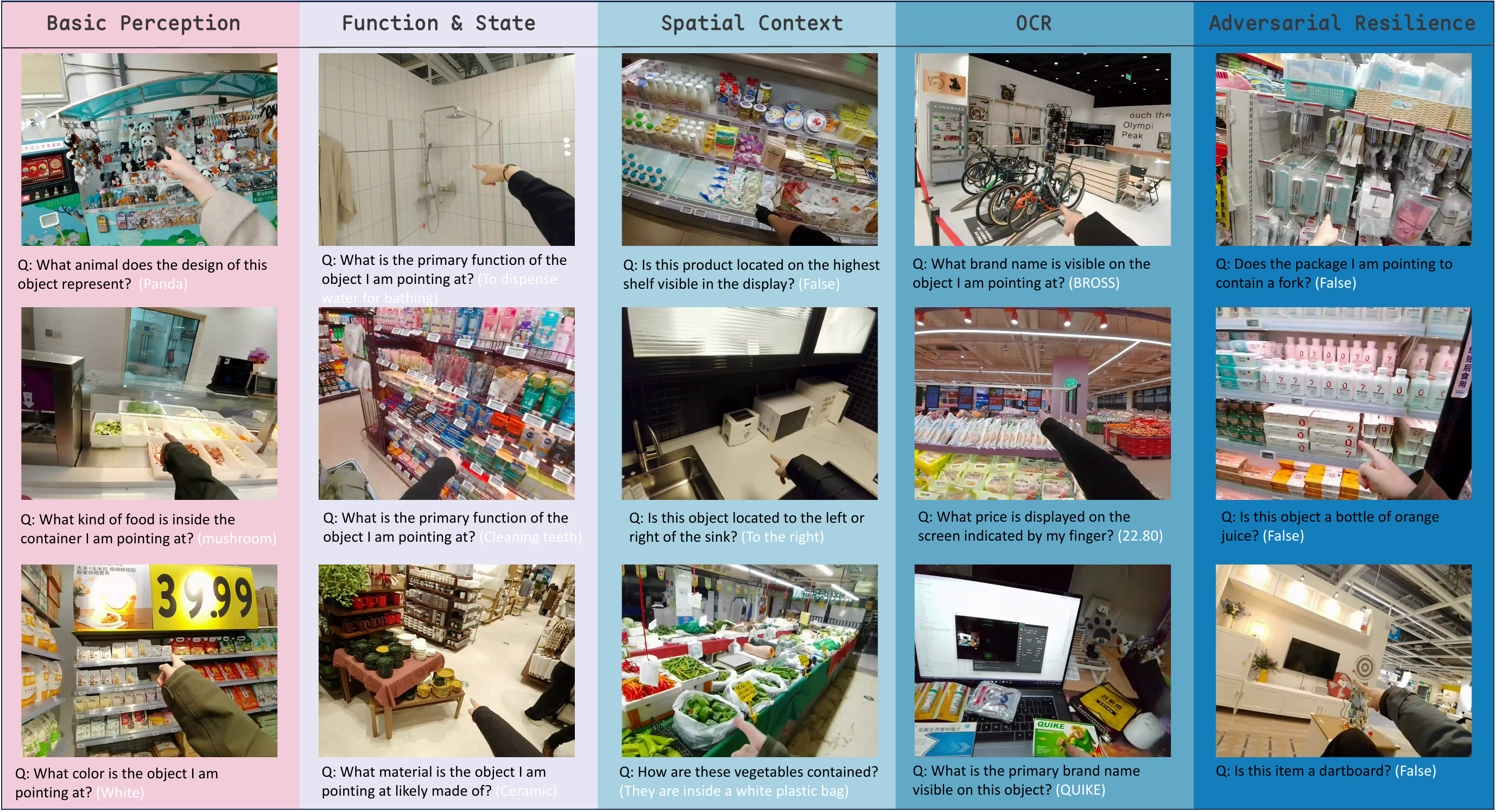}
    \caption{Qualitative success cases of the fine-tuned Qwen3-VL-8B across the five evaluation dimensions. For each dimension, we present three representative examples. The figure illustrates both the diversity of pointing-based questions in \textit{EgoPoint-Bench} and the fine-tuned model's ability to answer them correctly.}
    \label{fig:appendix_success_5dims}
\end{figure*}

\subsection{Success Cases Across Deixis Levels}
We further present representative examples across the three deixis levels: \textit{L1 (Explicit Action)}, \textit{L2 (Visual Locative)}, and \textit{L3 (Implicit Pronoun)}. In these examples, the original Qwen3-VL-8B fails to identify or reason about the pointed target, whereas the model fine-tuned on our synthetic dataset succeeds. These comparisons highlight that our method improves robustness under different levels of referential ambiguity.

\begin{figure*}[t]
    \centering
    \includegraphics[width=\textwidth]{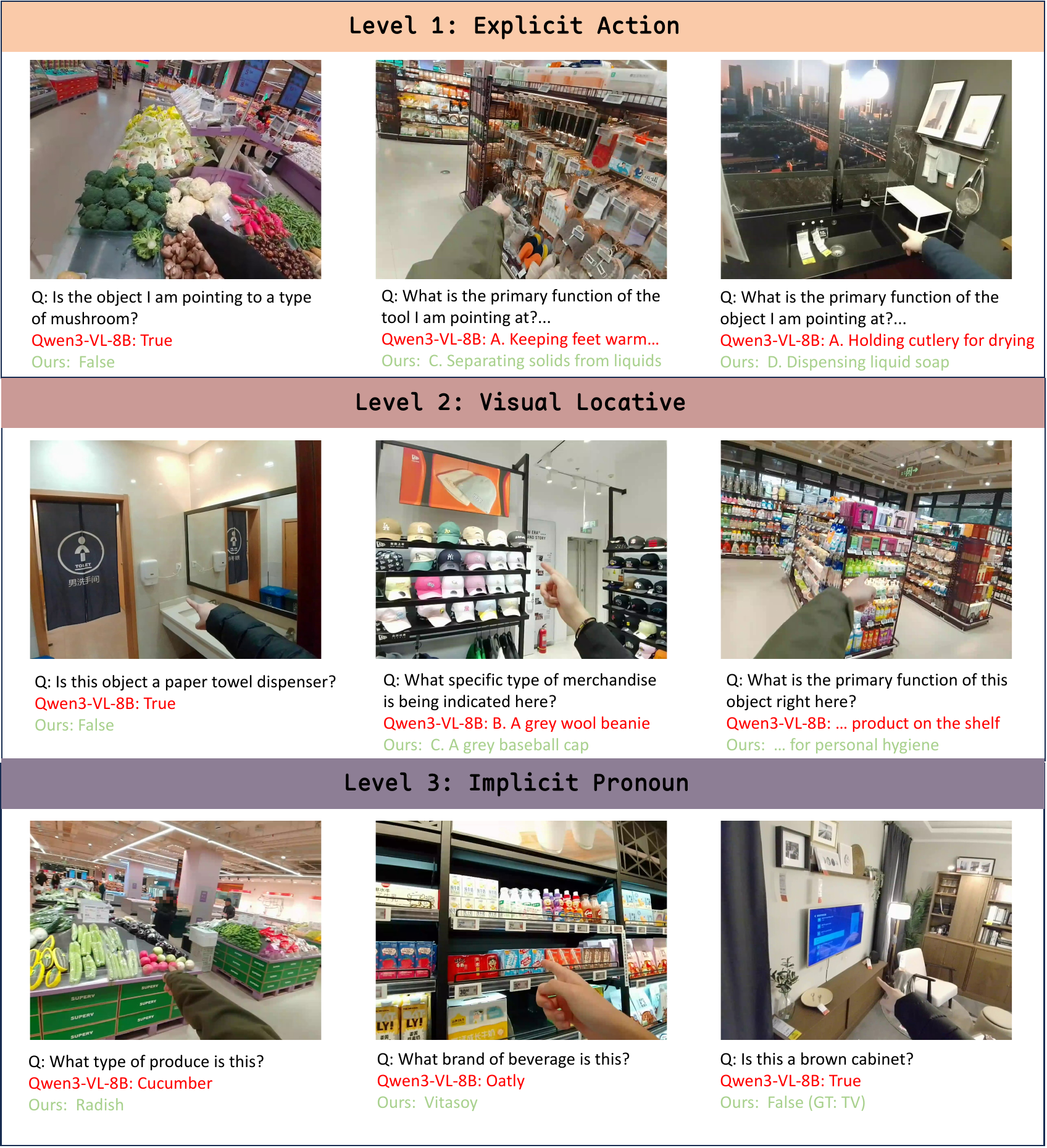}
    \caption{Representative comparison cases across the three deixis levels. In all examples, the original Qwen3-VL-8B fails, while the model fine-tuned on our synthetic data gives the correct answer. These cases demonstrate improved referential reasoning under explicit, locative, and highly implicit pointing expressions.}
    \label{fig:appendix_success_3levels}
\end{figure*}

\section{Additional Information}
\label{sec:appendix}

\subsection{Real-World Data Construction}
\label{subsec:real_world}

To bridge the domain gap between simulation and reality, we constructed a high-quality real-world dataset focusing on egocentric pointing interactions.

\subsubsection{Data Acquisition and Automated Pre-processing}
\label{subsubsec:acquisition}

\textbf{Automated Alignment Pipeline.} We designed a precision pipeline combining automated extraction with manual verification to achieve alignment across ``Pointing Action -- Target Object -- Speech Description -- Semantic QA.''

\begin{itemize}
    \item \textbf{Voice-Driven Keyframe Localization:} The process begins with speech recognition. We employed the industrial-grade open-source model \textbf{FunASR}\footnote{\url{https://github.com/modelscope/FunASR}} (\texttt{paraformer-zh}) to generate timestamped transcriptions.
    \begin{itemize}
        \item We defined a specific trigger word (e.g., ``Start'') to mark the onset of a pointing action.
        \item The system automatically detects the timestamp of this trigger and extracts the immediately following object noun as the candidate target.
        \item This process defines a temporal window of interest for visual extraction.
    \end{itemize}

    \item \textbf{Clarity-Aware Frame Selection:} To mitigate motion blur caused by head movements and device jitter, we implemented a \textbf{Multi-Metric Clarity Assessment} algorithm rather than random frame sampling. This algorithm fuses three complementary metrics:
    \begin{enumerate}
        \item \textbf{Laplacian Variance:} Captures high-frequency components to detect general focus blur.
        \item \textbf{Frequency Domain Analysis:} Analyzes the spectral energy distribution to identify motion blur patterns.
        \item \textbf{Edge Density:} Evaluates the sharpness of structural edges within the frame.
    \end{enumerate}
    By normalizing and computing a weighted fusion of these metrics (with all weighting coefficients set to 1.0), we assign a comprehensive clarity score to every frame within the identified time window. The top-performing frames with the highest scores are selected as candidate representative images.
\end{itemize}

\subsubsection{Human-in-the-Loop Annotation}
\label{subsubsec:annotation}

To ensure high quality, we employed a rigorous \textit{Human-in-the-Loop} (HITL) pipeline. The process involves close collaboration between annotators and data collectors to guarantee that annotations faithfully reflect the original pointing intent.

\textbf{Manual Annotation Workflow.} Based on the candidate clear frames selected by the automated algorithm, human annotators perform the following steps:
\begin{enumerate}
    \item \textbf{Frame Selection \& Privacy Protection:} Manually select the frames that clearly contain the hand gesture from the top candidates. Any visible faces in the background are blurred to protect privacy.
    \item \textbf{Transcription Verification:} Verify the correctness of the object name and description automatically transcribed by the ASR system.
    \item \textbf{BBox Annotation:} Manually draw Bounding Boxes (BBox) around the pointed-at object. This step requires deep cooperation and communication with the original data collectors to ensure the annotated object and BBox strictly align with the user's original pointing intention, especially in cluttered scenes. Each collector and annotator was paid \$15 per hour.
\end{enumerate}

\subsubsection{Environmental Diversity and Statistical Robustness}

To rigorously evaluate the Sim-to-Real generalization capabilities of MLLMs, our real-world dataset was explicitly curated to maximize environmental variance and ecological validity. Rather than relying on a single controlled laboratory setting, data collection spanned a wide array of unconstrained, dynamic environments. This high-diversity collection strategy ensures the benchmark effectively tests model robustness against background clutter, complex lighting variations, and unpredictable domain shifts.

We focused heavily on common daily life scenarios where users naturally rely on egocentric assistants for referential reasoning and information retrieval. The dataset comprises a highly diverse set of target instances distributed across various functional scenes:

\begin{itemize}
    \item \textbf{Retail and Groceries ($\approx$48\%):} Captured in supermarkets and fresh food markets, featuring densely packed items like produce (e.g., avocados, morels), snacks, and daily chemical products. These scenes introduce severe visual clutter, fine-grained occlusion, and challenging lighting.
    \item \textbf{Home and Furniture Environments ($\approx$35\%):} Collected in complex showrooms (e.g., IKEA) and apartments, encompassing furniture, home appliances, and kitchenware. These settings test spatial reasoning in environments with high structural coupling.
    \item \textbf{Apparel and Accessories ($\approx$5\%):} Recorded in clothing stores, involving items with high intra-class variance and textural ambiguity, such as hoodies, down jackets, and scarves.
    \item \textbf{Education, Sports, and Leisure ($\approx$5\%):} Covering interactions in classrooms and gyms, targeting items like stationery, basketballs, and dumbbells.
    \item \textbf{Public Infrastructure and Navigation ($\approx$4\%):} Focused on complex street-level interactions, such as pointing at traffic lights, crosswalks, public utilities, and vehicles in dynamic contexts.
    \item \textbf{Wildlife and Dynamic Subjects ($\approx$3\%):} Captured in zoos, introducing dynamic, non-rigid targets (e.g., pandas, lions, monkeys) against highly irregular natural backgrounds.
    \item \textbf{Healthcare and Pharmacy ($\approx$1\%):} Featuring safety-critical, highly specific items like medical supplies, disinfectants, and thermometers.
\end{itemize}

Across these environments, we collected hundreds of distinct fine-grained object categories. This extensive distribution yields a high category-to-sample ratio, ensuring that models cannot rely on memorized priors or spurious background correlations. 

From a statistical perspective, our real-world sample size is sufficiently large to yield a tight margin of error of approximately 3\% at a 95\% confidence interval. As demonstrated in Table \ref{tab:main_results_final}, the performance gap between our fine-tuned models and the base models ranges from roughly 5\% to 13\%. The observed gains of our fine-tuned models over their base counterparts are substantially larger than this scale, suggesting that the improvements are unlikely to be explained by sampling noise alone. The consistent performance gains achieved by models—which were trained exclusively on synthetic data—across these highly complex daily scenarios confirm that the "Point-Sim" pipeline successfully bridges the Sim-to-Real domain gap.

\subsection{QA Generation}
To synthesize QA pairs, Gemini 3 Pro is employed across our simulated and real-world datasets. We ensure the generation of high-fidelity labels by leveraging simulator-derived ground truth, specifically by superimposing red bounding boxes on the target objects. To further guide the model's reasoning, visual inputs are supplemented with exact object nomenclature and exhaustive descriptions. Regarding real-world samples, the original open-ended user queries are utilized as 
description for prompting. After manual validation, the refined prompt templates are formulated as follows:

\begin{tcolorbox}[
    breakable,
    enhanced,
    colback=gray!5,      
    colframe=gray!30,    
    boxrule=1pt,         
    arc=4mm,             
    title=\textbf{Data Generation Specialist Prompt}, 
    coltitle=black,      
    colbacktitle=gray!10,
    fontupper=\ttfamily\small, 
    left=6mm, right=6mm, top=4mm, bottom=4mm 
]

    {\color{blue!70!black}\textbf{SYSTEM\_PROMPT}} \\
    \# Role \\
    You are an expert Data Generation Specialist for Vision-Language Models. \\
    Your goal is to create ONE single, high-quality Question-Answer pair for an egocentric image based strictly on the specific constraints provided by the user. \\
    
    \vspace{0.2cm}
    \# Context \\
    You will be provided with: \\
    1. The \textbf{Target Object} name (Ground Truth). \\
    2. The \textbf{Target Object} description or question. \\
    3. The specific \textbf{Dimension} (e.g., Affordance, Basic Perception). \\
    4. The specific \textbf{Deixis Level} (how the object is referenced). \\
    5. The specific \textbf{Question Type} (e.g., Multiple Choice). \\
    
    \vspace{0.2cm}
    {\color{red!70!black}\# Critical Constraint: The ``Red Box'' Rule} \\
    - The target object is highlighted with a red bounding box in your internal vision. \\
    - \textbf{NEVER} mention ``red box'', ``rectangle'', ``highlight'', or ``outline'' in the text. \\
    - Pretend the user is pointing at the object with their finger. \\
    
    \vspace{0.2cm}
    \# Guidelines for Quality \\
    \#\# 1. Anti-Cheating Option Generation (Crucial for Multiple Choice) \\
    You must avoid ``lazy'' distractors. Follow this logic to generate options: \\
    - \textbf{Correct Answer:} The ground truth label or attribute. \\
    - \textbf{Distractor 1 (Scene Hard Negative):} An object that is \textbf{present elsewhere in the image} but NOT being pointed at. \\
    - \textbf{Distractor 2 (Visual Hard Negative):} An object sharing similar \textbf{color, shape, or texture} with the target. \\
    - \textbf{Distractor 3 (Contextual Hard Negative):} An object plausibly found in this specific environment, but definitely NOT the target. \\
    - \textbf{Verification:} Ensure the correct answer is unique and unambiguous among options. \\
    
    \vspace{0.2cm}
    \#\# 2. Zero-Leakage Question Formulation \\
    - \textbf{The ``Blindfold'' Test:} If a human can guess the answer just by reading the question (without the image), the question is BAD. \\
    - \textbf{Bad:} ``What is this red round fruit?'' (Reveals color, shape, category). \\
    - \textbf{Good:} ``What is the name of this object?'' (Reveals nothing). \\
    
    \vspace{0.2cm}
    {\color{purple!80!black}\# Definitions of Constraints} \\
    \#\# Deixis Levels (Reference Style) \\
    - \textbf{L1 (Explicit Action):} ``the object I am pointing at'', ``what is indicated by my finger''. \\
    - \textbf{L2 (Visual Locative):} ``this object right here'', ``that thing over there''. \\
    - \textbf{L3 (Implicit Pronoun):} ``this'', ``it'', ``this one''. \\
    
    \vspace{0.2cm}
    \#\# Dimensions (Question Topic) \\
    - \textbf{Basic Perception:} category, color, shape, material, counting. \\
    - \textbf{Affordance \& Function:} Edibility, operation method, state, safety, utility. \\
    - \textbf{Context \& Relation:} Spatial position, scene compatibility. \\
    - \textbf{OCR \& Text:} Reading text on the object. \\
    - \textbf{Adversarial:} Asking about non-existent parts or false premises. \\
    
    \vspace{0.2cm}
    \#\# Question Types \\
    - \textbf{True\_False:} Answer is ``True'' or ``False''. \\
    - \textbf{Multiple\_Choice:} Provide 4 options (A/B/C/D). \\
    - \textbf{Open\_Ended:} Answer is a concise phrase. \\
    
    \vspace{0.2cm}
    {\color{orange!80!black}\# Output Format} \\
    Output \textbf{ONLY} a pure JSON object containing the single generated pair. \\
    \textbf{JSON Structure:} \\
    \{ \\
    \hspace*{1em}``qa\_pairs'': [ \\
    \hspace*{2em}\{ \\
    \hspace*{3em}``question'': ``string'', \\
    \hspace*{3em}``options'': [``A. string'', ``B. string'', ``C. string'', ``D. string''] OR null, \\
    \hspace*{3em}``answer'': ``string'', \\
    \hspace*{3em}``dimension'': ``string'', \\
    \hspace*{3em}``deixis\_level'': ``string'', \\
    \hspace*{3em}``type'': ``string'', \\
    \hspace*{3em}``rationale'': ``string'' \\
    \hspace*{2em}\} \\
    \hspace*{1em}] \\
    \} \\

    \vspace{0.6cm}
    \hrule
    \vspace{0.4cm}

    {\color{blue!70!black}\textbf{USER}} \\
    I need you to generate a QA pair for the following object based on these strict requirements: \\
    
    \vspace{0.2cm}
    1. \textbf{Target Object (Ground Truth):} \promptvar{\{obj\_name\}} \\
    2. \textbf{Description or Question:} \promptvar{\{description\}} \\
    3. \textbf{Required Dimension:} \promptvar{\{dimension\}} \\
    4. \textbf{Required Deixis Level:} \promptvar{\{deixis\_level\}} \\
    5. \textbf{Required Question Type:} \promptvar{\{q\_type\}} \\
    
    \vspace{0.2cm}
    \textbf{Instruction:} Generate a question that strictly fits the dimension above. \\
    Use the specified deixis phrasing style. \\
    Format the answer according to the question type. \\
    Ensure no leakage of the object's name in the question.

\end{tcolorbox}

\end{document}